\pdfoutput=1

\RequirePackage{fix-cm}
\RequirePackage{rotating}%
\documentclass[smallextended]{svjour3}       
\smartqed 
\usepackage{cite}
\usepackage{mathptmx}
\usepackage{amsmath,amssymb,amsfonts}
\usepackage{graphicx}
\usepackage{subcaption}
\usepackage{textcomp}
\usepackage{xcolor}
\usepackage{listings}
\usepackage{multirow}
\usepackage{wrapfig}
\usepackage{xcolor,colortbl}
\usepackage[flushleft]{threeparttable}
\usepackage{amsmath}
\usepackage{relsize}
\usepackage{pifont}
\usepackage[sort&compress,numbers]{natbib}
\usepackage[skins]{tcolorbox}
\usepackage{wasysym}
\usepackage{pgfplots}
\usepackage{changepage} 
\usepackage{enumerate}
\usetikzlibrary{shadows}
\usepackage{enumitem}
\usepackage[ruled,vlined]{algorithm2e}
\usepackage{adjustbox,booktabs,multirow}

\usepackage{algorithmicx}
\usepackage[hidelinks]{hyperref}
\usepackage{tcolorbox}
\newtcolorbox{blockquote}{colback=gray!5,boxrule=0.4pt,colframe=black,fonttitle=\bfseries}
\usepackage[T1]{fontenc}

\usepackage{cite}

\newcommand{\bi}{\begin{itemize}}
\newcommand{\ei}{\end{itemize}}
\newcommand{\be}{\begin{enumerate}}
\newcommand{\ee}{\end{enumerate}}

\usepackage{subcaption,siunitx,booktabs}

\pgfplotsset{compat=1.10}
\usepgfplotslibrary{fillbetween}

\def\BibTeX{{\rm B\kern-.05em{\sc i\kern-.025em b}\kern-.08em
    T\kern-.1667em\lower.7ex\hbox{E}\kern-.125emX}}
    
\definecolor{MyDarkBlue}{rgb}{0,0.08,0.45} 
\newcommand\MyBox[2]{
  \fbox{\lower0.75cm
    \vbox to 1.7cm{\vfil
      \hbox to 1.7cm{\hfil\parbox{1.4cm}{#1\\#2}\hfil}
      \vfil}%
  }%
}

\newcommand{\IT}{\sffamily{GroupCART}}

\journalname{Empirical Software Engineering}
\begin{document}

\title{Software Engineering Principles for Fairer Systems: Experiments with {\IT}}

\author{Kewen Peng         \and
        Hao Zhuo           \and
        Yicheng Yang       \and
        Tim~Menzies
}

\institute{Kewen Peng,  Tim Menzies\\
Department of Computer Science, North Carolina State University, Raleigh, USA. \email{kpeng@ncsu.edu, timm@ieee.org}\\
\newline
Hao Zhuo, Yicheng Yang\\
Cornell University, Ithaca, USA. \email{hz324@cornell.edu, yy546@cornell.edu}
}

\date{Received: date / Accepted: date}

\maketitle

\begin{abstract}
\small
{\em Context:} Discrimination-aware classification aims to make accurate classifications with certain fairness constraints. 
Traditional machine learning software based on decision tree learners takes into account only the information gain change in the target attribute.
Although effective, such learners generate models
that inappropriately discriminate against specific protected social groups (e.g., gender, ethnicity, etc.).
Motivated by such results, more recent studies have proposed bias mitigation methods.

{\em Objective}: Rather than repair the bias once after it has been added to a model, we take another approach, and we {\em avoid bias during
model generation}. Specifically, during decision tree construction, we make choices that better navigate the fairness-performance trade-off.

{\em Method:} Here, we propose {\IT}, a tree-based ensemble optimizer that is designed not only to optimize for the {\em decrease in entropy} in the target attribute but also the {\em entropy increase}  in the protected attributes.

{\em Results:} The experiments of this paper show that, without transforming the data, we can train a more fair decision tree model with a negligible change in model performance. Further, this method can be tuned with customized weights: we can offer a smooth and flexible trade-off between performance and fairness depending on potential user requirements. 

{\em Conclusion:} Algorithmic bias within decision tree models can be mitigated via multi-task fairness-aware learning. As evidence for this, 
our proposed {\IT} can provide an optimal performance-fairness trade-off compared to previous works.

{\em Reproduction Package:} To better support open science, all scripts and data used in this study are available online at \url{https://github.com/anonymous12138/groupCART}.

\end{abstract}

\section*{Conflict of interest}
The authors declare that they have no conflict of interest.

\newpage

\section{Introduction}
\label{intro}

In recent years, machine learning (ML) has emerged as a transformative technology with profound implications in software engineering (SE). Its ability to analyze large amounts of data and make predictions has led to remarkable advances in fields such as healthcare, finance, and autonomous systems. However, as machine learning algorithms increasingly shape decision-making processes, concerns about fairness and ethical implications have garnered significant attention. The quest for fairness in machine learning, guided by the principles of AI ethics, has become a crucial endeavor to ensure equitable outcomes, address biases, and uphold societal values.

The motivation for using fairness-aware ML in software engineering is two-fold. First, as ML algorithms become more prevalent in our society, it is critical to ensure that they are not perpetuating and reinforcing existing societal biases and discrimination. For example, if an algorithm is trained on historical data that are biased against certain groups, it may make discriminative decisions against those groups. This can lead to unfair treatment of individuals and perpetuate existing social inequalities. By incorporating fairness considerations into the design and implementation of ML algorithms, we can reduce the risk of bias and discrimination.
Second, there is also a legal and ethical imperative to maintain a fair use of ML techniques. Discrimination on the basis of protected characteristics (race, gender, etc.) is prohibited by law in many countries, and failure to take steps to prevent discrimination can result in legal and reputational consequences. 

Many empirical studies in software engineering show that mitigating fairness bugs is associated with a trade-off between fairness and performance: To obtain fairer outcomes from a machine learning model, one must compromise the predictive power (accuracy, etc.) of the model.
However, in more recent studies, researchers have reported encouraging progress that such trade-offs can be optimized \cite{cruz2021promoting,tizpaz2022fairness,hort2024search}. \begin{wrapfigure}{r}{2.0in}
\begin{adjustbox}{max width=.99\textwidth}
\includegraphics[width=.99\linewidth]{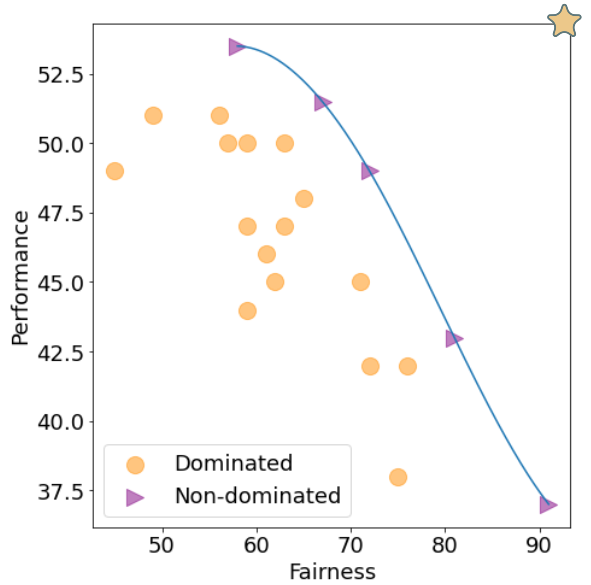}
\end{adjustbox}
\caption{Pareto frontier in the Adult dataset with protected attribute {\it race}.}
\end{wrapfigure}\label{pareto}
That is to say, nowadays more research is conducted aiming to ensure fairness with the least harm to model performance. Among various approaches, Ensemble learning is one of the most explored techniques.
State-of-the-art (SOTA) approaches, such as Fair-SMOTE \cite{Chakraborty2021BiasIM} and MAAT \cite{chen2022maat}, use different forms of ensemble learning to achieve an ideal trade-off balance between performance and fairness. However, we discover that little effort was invested in prior work to conduct a comprehensive assessment of the trade-off goals space. In other words, in this paper, we ask the following question:
\begin{enumerate}
    \item Can we program "awareness of fairness" as a {\bf hyperparameter configuration} within the ML model?
    \item Can we optimize such fairness-aware configurations as a {\bf multi-objective optimization} task where there exist trade-offs between predictive performance and fairness?
    \item If one can obtain different optimum trade-offs (with different model configurations),  could we aggregate these potential solutions via {\bf ensemble learning}? 
\end{enumerate}

Previous work such as MAAT \cite{chen2022maat} failed to answer those questions as their approach only explored the ensemble learning of size two: one model for fairness and one model for accuracy. 
We wonder if more than two models can obtain better performance-fairness trade-offs. Moreover, if users of ML software would like to use different fairness or performance metrics as their key measurements, we would like to make our algorithm customizable to different metrics, as well as to different importance ratios between fairness metric scores and performance scores.

\begin{figure}[b!]
\centering
\includegraphics[width=.75\linewidth]{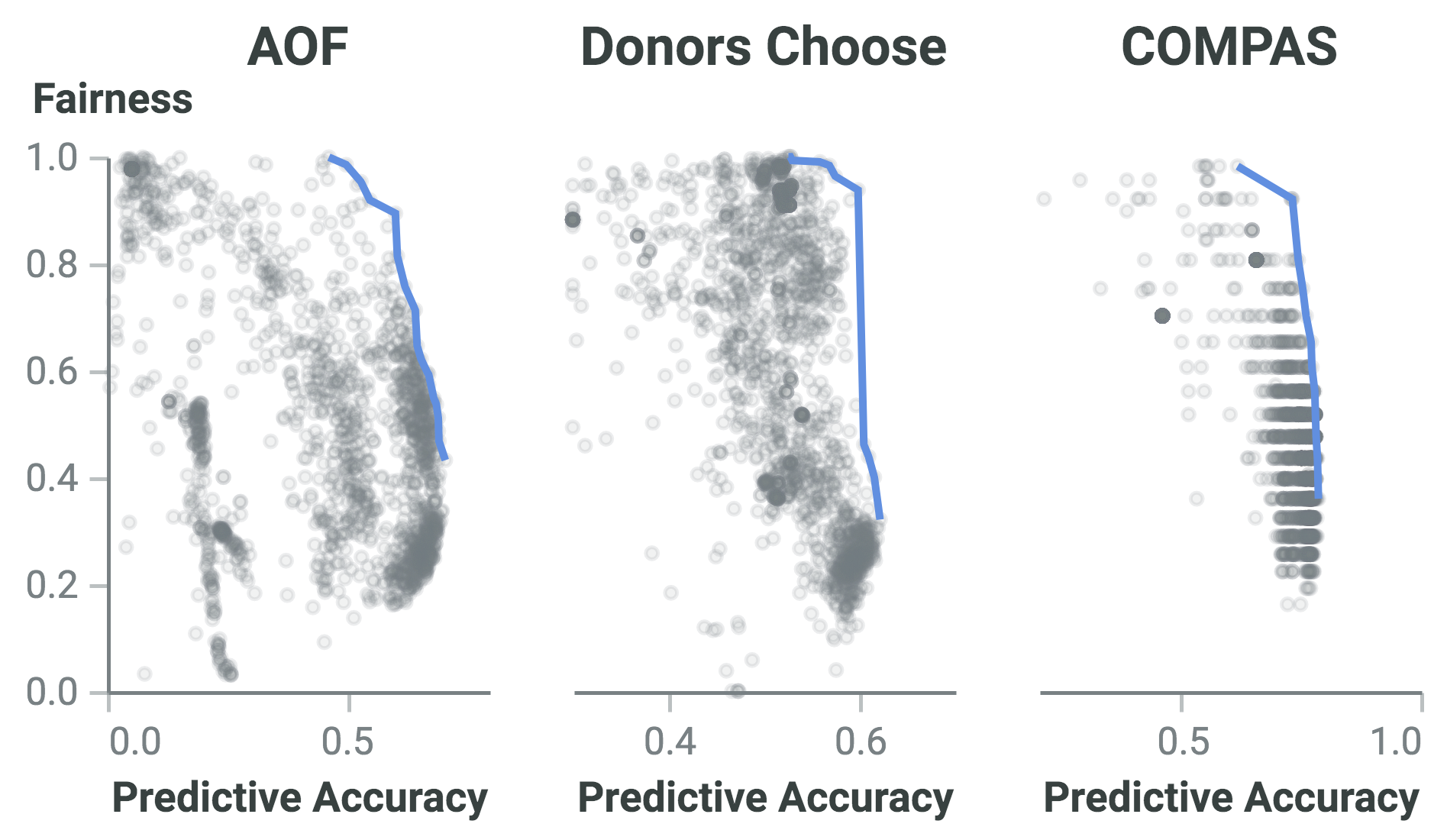}
\caption{Presented in the paper be Cruz et al. \cite{cruz2021promoting}, the Pareto frontier of the fairness-performance trade-off in three datasets. Each dot represents a hyper-parameter configuration setting applied on one of the five ML algorithms selected by Cruz et al..} 
\end{figure}\label{cruz}

Preliminary results in this paper demonstrate that, by training candidate models with different configuration parameters, one can obtain more than one ``optimal trade-off" solution. As shown in Figure~\ref{pareto}, the x-axis represents disparate impact (DI) scores, and the y-axis the F1 scores. In both axes, greater values are better. Non-dominated points are calculated according to the definition of binary domination as described later in Eq.~\ref{b-dom}. The star at the top-right corner represents the ``heaven" where both fairness and performance reach optimum.
We initialized 20 decision tree models, each with a distinct importance weight ratio between performance and fairness. Based on the definition of binary domination, there exist 5 non-dominated (so-called Pareto optimal) models. More importantly, this leads to a crucial question: Since the 5 models are all considered ``optimal'' solutions in the performance-fairness trade-off, which model(s) should we choose to generate final prediction outcomes? A similar phenomenon is observed in the study by Cruz et al. \cite{cruz2021promoting}.  In their recent study, the fairness-performance Pareto frontiers were obtained from training 10K different hyperparameter configurations of 5 ML algorithms. As shown in Figure 2, most of the configurations are, in fact, suboptimal solutions in the fairness-performance multi-objective competition. In this paper, we found that we can endorse the observation made by Cruz et al. in many other empirical datasets. More importantly, our proposed approach can reduce the size of the configuration pool from thousands to tens of candidates, as we directly configure the awareness of fairness via fairness-aware decision tree algorithms (later elaborated in \S\ref{fdt}). 

In this paper, we propose an ensemble learning framework that (a) trains decision trees that simultaneously optimize for both fairness and predictive performance, (b) identifies distinct optimal solutions as candidate trees, and (c) averages the predictions of the candidate trees to make the final predictions. As shown by the experimental results, {\IT} has achieved the following contributions:
\begin{itemize}
    \item {\IT} provides on-par or better predictive performance than benchmark methods. 
    \item {\IT} provides adjustable tuning options for trade-offs between performance and fairness.
    \item {\IT} is applicable to multiple protected attributes and can provide smooth tuning options.
    \item The implementation of {\IT} is publicly available and potentially applicable to any tree-based ensemble learning algorithm (e.g. Random Forest, AdaBoost decision tree).
\end{itemize}

\section{Background}
\subsection{Why SE Should Care About Fairness}

The rapid development of ML has greatly benefited SE practitioners, and examples of ML-assisted software can be found everywhere: defect prediction models used to locate the most error-prone code files in the upcoming releases; effort estimation tools used to better manage human and capital resources; multi-objective optimizers used to generate configuration solutions for systems of enormous configurable options. Meanwhile, ethical concerns have also drawn increasing attention in the ML and SE communities. 

While in many scenarios, the only utility needed to be optimized is the performance of the models (in tasks about prediction, classification, ranking, etc.), other cases where private information of human beings is collected need to be handled more carefully. ML software systems have been deployed in many areas to assist in making decisions that affect human individuals: courts and corrections departments in the US use software to determine sentence length for defendants~\cite{feller2016computer}; algorithms are used to predict default payments from credit card users~\cite{yeh2009comparisons}. During such procedures, private information, such as age, ethnicity, and gender, is collected.
Moreover, previous studies reported that models learned from such data may contain algorithmic bias toward certain social groups.

In response to the above-mentioned issues, IEEE has provoked ethical designs of AI-assisted systems~\cite{shahriari2017ieee} and the European Union also announced the ethics guidelines for building trustworthy AI~\cite{doi/10.2759/177365}. Fairness has been emphasized in both documents. Big industrial companies such as Facebook~\cite{Fairness_Flow}, Microsoft~\cite{FATE}, and Google~\cite{simonite2020google} have also begun to invest effort in ensuring the fairness of their products. In academia, IEEE and ACM have set specific tracks~\cite{FAT,EXPLAIN} for papers studying fairness problems.

\subsection{Related Work: Fairness in SE}
Despite everything mentioned above, sometimes we get questions about the relevance and applicability of this work to SE.
For example: ``Given a paper proposing methods to fix fairness bugs in ML-assisted software, should one submit this paper instead to a machine learning venue?''. Here, we demonstrate the relevance of ML fairness in the SE field from the following perspectives.

\begin{itemize}
\item {\bf Software Configuration}:
Recent research has increasingly approached fairness in ML as a problem of software configuration~\cite{Chakraborty2021BiasIM,cruz2021promoting,peng2022fairmask,peng2023veer,nair2018finding}, offering compelling evidence that fairness can be integrated into the software development life cycle through systematic interventions. 
One significant study~\cite{Chakraborty_2020} argues for the routine inclusion of bias testing and mitigation within the ML software development process, demonstrating that these practices can enhance fairness without significantly compromising model performance. 
Complementing this, another study~\cite{chen2022maat} introduces a novel ensemble approach, MAAT, which addresses both fairness and performance bugs, highlighting the potential of ensembles in this domain. 
Further work~\cite{gohar2023towards} on ensemble methods explores how fairness can be inherently improved by carefully designing the ensemble architecture without relying on traditional mitigation techniques, suggesting that fairness composition can be managed as part of software configuration based on data characteristics.

\item {\bf Feature Engineering and Debugging}:
Other approaches focus on more granular aspects of the software development process, such as feature engineering and debugging, to enhance fairness. 
For example, enlarging the feature set~\cite{zhang2021ignorance} can strike a balance between fairness and accuracy, as highlighted in research that emphasizes the importance of richer feature sets for achieving this sweet spot. 
Moreover, many researchers emphasize the importance of treating fairness issues similarly to software bugs. A novel debugging method~\cite{li2022training} referred to as Linear-regression based Training Data Debugging (LTDD), inspired by traditional software debugging practices, is proposed to identify and exclude biased feature components in training data, thereby improving fairness. 
Similarly, an automatic repair approach~\cite{hort2024search} leverages multi-objective search techniques to enhance both fairness and accuracy in ML-based decision-making software, reinforcing the idea that fairness can be optimized through software configuration. 
Additionally, studies on counterfactual predictions, such as MirrorFair~\cite{xiao2024mirrorfair}, showcase how fairness bugs can be mitigated by training models on both original and counterfactual datasets, further underscoring the potential of systematic configurations in achieving fairer ML outcomes. 
These works collectively suggest that fairness in ML can be effectively managed through targeted software configuration strategies, paving the way for more equitable and reliable software systems.

\item {\bf Fairness Testing}:
Fairness testing is another critical area where SE practices can be adapted to address bias in ML systems. Traditional testing methods have been expanded to include fairness as a key criterion, ensuring that models are evaluated not only for performance but also for equity. For example, adversarial sampling techniques have been developed to rigorously test ML models for fairness~\cite{zhang2020white}, helping to identify and mitigate potential biases before deployment. 
Besides explorations in specific techniques, a comprehensive survey~\cite{chen2024fairness} of 100 fairness testing papers has further highlighted the importance of this practice within SE, demonstrating how systematic testing can prevent the propagation of biases in ML-assisted software. It compiles a summary of public datasets and open-source tools for fairness testing, providing a navigation for researchers and practitioners interested in the field.
These contributions emphasize the proactive role of fairness testing in the software development process, making it an integral part of creating unbiased software systems.

\item {\bf Empirical Studies}:
Despite the existing body of work in the field, the persistent challenge of balancing fairness and performance in machine learning models underscores the complexity of treating fairness as a software engineering problem. 
As mentioned earlier, it's commonly believed in the research community that bias mitigation techniques tend to enhance fairness, but often at the expense of machine learning performance (such as accuracy), a concept referred to as the "fairness-performance trade-off." ~\cite{chen2022maat,wick2019unlocking}
A comprehensive empirical study~\cite{chen2023comprehensive} investigating 17 bias mitigation methods across various machine learning classifiers reveals the intricate trade-offs involved in this process. 
The study highlights that the effectiveness of these methods is highly context-dependent, varying across tasks, models, and the choice of protected attributes. No single bias mitigation method consistently achieves the optimal trade-off between fairness and performance across all scenarios, emphasizing the need for software engineers to carefully select and configure these methods based on specific application requirements. 
This evidence reinforces the notion that fairness must be treated as a configurable software property, requiring thoughtful integration into the software development life cycle.
\end{itemize}

\section{Problem Description}\label{rw}
In this section, we introduce (a) fundamental concepts of  software fairness and 
(b) related work that tries to ensure it. 

\subsection{Fairness in ML Software}
In this work, we use binary classification models. We define some terms specific to the fairness of binary classification. 
\begin{itemize}
    \item A {\em favorable label} in a binary classification task is the label that grants the instance (usually human individuals) with privilege such as a job offer or being accepted for a loan.
    \item A {\em protected / sensitive attribute} reveals the social groups to which data instances belong, such as gender, race, and age. A binary protected attribute will divide the entire population into {\em privileged} and {\em unprivileged} groups in terms of the difference in receiving the favorable label. 
\end{itemize}
The notion of bias comes if the outcome of the classification model is significantly affected by sensitive/protected attributes. Table~\ref{tab:dataset} shows seven fairness datasets used in this work. These datasets are very popular in the fairness domain and have been used by many previous researchers~\cite{Chakraborty_2020,Chakraborty2021BiasIM,chakraborty2019software,9286091,Biswas_2020}. All of these data sets contain at least one protected attribute. Depending on that, the population is divided into two groups, which get different benefits. For example, in the Adult~\cite{ADULT} dataset, there are two protected attributes. Based on ``sex'', ``male'' is privileged; Based on ``race'', ``white'' is privileged.

The concept of fairness is complicated and very domain-specific. Narayanan~\cite{Arvind} has defined 21 different versions of fairness. Based on prior literature~\cite{Biswas_2020,Chakraborty_2020,Chakraborty2021BiasIM}, among these 21 versions, two specific versions of fairness are widely explored and given the greatest importance. We decided to explore the same two versions and chose different metrics to evaluate them. 

\begin{itemize}
    \item {\em Group fairness} requires the approximate equalization
    of certain statistical properties across groups divided by the protected attribute. In this paper, we use 4 group fairness metrics that were widely used in previous research~\cite{kamiran2012data,feldman2015certifying,Chakraborty_2020,Chakraborty2021BiasIM,Biswas_2020}.
    \item {\em Individual fairness} requires that similar individuals receive similar prediction outcomes using the ML model. The usual metric for measuring individual fairness is ``consistency''. But ``consistency'' is a collective metric based on nearest neighbors. This means that it can be calculated for a set of data points, not for a single point. Chakraborty et al.~\cite{Chakraborty_2020} came up with a new metric for measuring individual fairness, where they measured the FLIP rate which computed the ratio of the population whose prediction outcomes are flipped (e.g., accepted to rejected) when reversing their protected attributes. We decided to use the same metric called {\it Flip Rate} (FR).
\end{itemize}
Table~\ref{tab:metrics} contains mathematical definitions of 5 fairness metrics. All group fairness metrics are calculated based on the binary classification confusion matrix, which consists of four parts:
true positive (TP), true negative (TN), false positive (FP), and false negative (FN).

\begin{table*}[t!]

\centering
\footnotesize
\caption{Description of datasets used in this paper.}
\resizebox{\textwidth}{!}{%
\begin{tabular}{cccccc}
\toprule
Dataset &
  \#Features &
  \#Rows &
  \multicolumn{1}{c}{Domain} &
  Protected Attribute &
  Favorable Label \\
  \midrule
Adult Census~\cite{ADULT} & 14  & 48,842 & U.S. census information from 1994 to predict personal income & Sex, Race &
  Income $>$ \$50,000 \\
Compas~\cite{COMPAS}         & 28    & 7,214  & Criminal history of defendants to predict reoffending & Sex, Race & Re-offend $=$ false   \\
German Credit~\cite{GERMAN}  & 20    & 1,000  & Personal information to predict good or bad credit                              & Sex       & Credit $=$ good   \\
Bank Marketing~\cite{BANK} & 16 & 45,211 & Marketing data of a Portuguese bank to predict term deposit                     & Age       & Subscription $=$ yes \\
Heart Health~\cite{HEART}   & 14    & 297    & Patient information from Cleveland DB to predict heart disease                  & Age       & Diagnose $=$ yes   \\
MEPS15~\cite{MEPS}        & 1831  & 4,870 & Surveys of household members and their medical providers & Race      & Utilization $>=$ 10  \\
\bottomrule
\end{tabular}%
}
\label{tab:dataset}
\end{table*}

\begin{table*}[t!]
\centering
\scriptsize
\caption{Definitions and descriptions of fairness metrics used in this paper.}
\resizebox{\textwidth}{!}{%
\begin{tabular}{|l|l|l|}
\hline
\multicolumn{1}{|c}{Metric} &
  \multicolumn{1}{|c|}{Definition} &
 \multicolumn{1}{c|}{Description} \\ \hline
Average Odds Difference (AOD) &
  \begin{tabular}[c]{@{}l@{}}TPR $=$ TP/(TP + FN), FPR $=$ FP/(FP + TN)\\ AOD= (($FPR_{U} - FPR_{P}$) + ($TPR_{U} - TPR_{P}$))/2\end{tabular} &
  \begin{tabular}[c]{@{}l@{}}Average of difference in False Positive Rates(FPR) and True\\ Positive Rates(TPR) for unprivileged and privileged groups\end{tabular} \\ \hline
Equal Opportunity Difference (EOD) &
  EOD $=$ $TPR_{U} - TPR_{P}$ &
  \begin{tabular}[c]{@{}l@{}}Difference of True Positive Rates(TPR) for unprivileged and\\ privileged groups\end{tabular} \\ \hline
Statistical Parity Difference (SPD) &
  SPD $=$ P (Y $=$ 1$|$PA $=$ 0) $-$ P (Y $=$ 1$|$PA $=$ 1) &
  \begin{tabular}[c]{@{}l@{}}Difference between probability of unprivileged group \\ (protected attribute PA $=$ 0) gets favorable prediction (Y $=$ 1)\\ \& probability of privileged group (protected attribute PA $=$ 1)\\ gets favorable prediction (Y $=$ 1)\end{tabular} \\ \hline
Disparate Impact (DI) &
  DI $=$ P {(}Y $=$ 1$|$PA $=$ 0{]}$/$P {[}Y $=$ 1$|$PA $=$ 1{)} &
  Similar to SPD but measuring ratio rather than the probability \\ \hline
Flip Rate (FR) &
  FR $=$ $\Sigma$(L$|$L{[}PA$=$0{]} $\not =$ L{[}PA$=$1{]}) / $total$ &
  \begin{tabular}[c]{@{}l@{}}The ratio of instances whose predicted label ($L$) will change\\ when flipping their protected attributes (e.g., PA$=$1 to PA$=$0) \end{tabular} \\ \hline
\end{tabular}%
}
\label{tab:metrics}
\end{table*}

\subsection{Bias Mitigation}\label{bias-mitigation}
Many researchers endeavor to ensure
fairness within their AI decision-making software. In the literature, bias mitigation methods can be categorized into three major groups, depending on when the mitigation procedure is performed.\\
    \noindent
    {\bf Pre-processing}: Pre-processing algorithms attempt to mitigate the bias of the model by pre-processing the training data that the model learns from. Reweighing was proposed by Kamiran et al.~\cite{kamiran2012data} to learn a probabilistic threshold that can generate weights to different instances in training samples according to the combination (protected and class attributes) that each of them belongs to. Fair-SMOTE~\cite{Chakraborty2021BiasIM} proposed by Chakraborty et al.~\cite{Chakraborty2021BiasIM} resamples and generates synthetic instances among the training data so that the training data can reach equal distributions not only between different target labels but also among different protected attributes.\\
    \noindent
    {\bf In-processing}: In-processing methods generally take the optimization approach to mitigate bias. The data set is typically divided into three parts: training, validation, and testing. The learner is fitted to the training set and then optimized to the validation set using both performance and fairness metrics as objectives. Exponentiated gradient reduction (EGR) is an in-processing technique that reduces fair classification to a sequence of cost-sensitive classification problems. Given the specific fairness metrics selected by users, EGR will return a randomized model with the lowest empirical error subject to the corresponding constraints \cite{agarwal2018reductions}.\\
    \noindent
    {\bf Post-processing}: This approach believes that bias can be removed by identifying and then reversing biased outcomes from the classification model. Thus, such methods typically only mutate the outcomes of the classification model, rather than the model itself. Kamiran et al. ~\cite{Kamiran:2018:ERO:3165328.3165686} proposed a ``reject option classification'' approach~\cite{Kamiran:2018:ERO:3165328.3165686} to first identify the model decision boundary with the highest uncertainty. Within that region, the method will adjust the ratio between favorable labels on unprivileged groups and unfavorable labels on privileged groups.\\

In particular, the above three genres are defined based on the phase in which an algorithm intervenes to mitigate bias. There are also many methods that cannot be explicitly categorized into a single genre. For example, the FairMask framework proposed by Peng et al. \cite{peng2022fairmask} in 2022 generates synthesized protected attributes (PAs) parallel to the model training phase, and synthetic attributes are used to replace real protected attributes during inference time. FairMask could be considered a post-processing method since it mitigates bias by identifying and reversing biased model outcomes via the use of synthetic PAs. However, the synthesis procedure occurs before/during the model-training phase. Therefore, it is difficult to clearly categorize this algorithm into any of the three types.

\section{Methodology} 
This section elaborates on the methodologies adapted in our proposed framework and previous studies in these domains. Moreover, we present our algorithmic designs for GroupCART as well as the intuition behind them.
\subsection{Fairness-aware Decision Tree (FDT)} \label{fdt}
The Fairness-aware Decision Tree (FDT) refers to a genre of tree-based algorithms that aim to address fairness concerns in decision-making processes \cite{kamiran2010discrimination,aghaei2019learning,zhang2019faht,ranzato2021fairness}. It extends the traditional decision tree algorithm by incorporating fairness constraints or modifications to ensure equitable outcomes across different groups.
Traditional decision tree learners use the Information Gain (IG) as the sole splitting criterion when constructing the subleaves. Fairness-aware decision trees, on the other hand, combine multiple constraints as the splitting criterion. Studied in prior works, FDT computes not only the IG on the target attribute for each potential split but also the IG on the protected/sensitive attributes.

\begin{figure*}[t!]
\centering
\includegraphics[width=.99\textwidth]{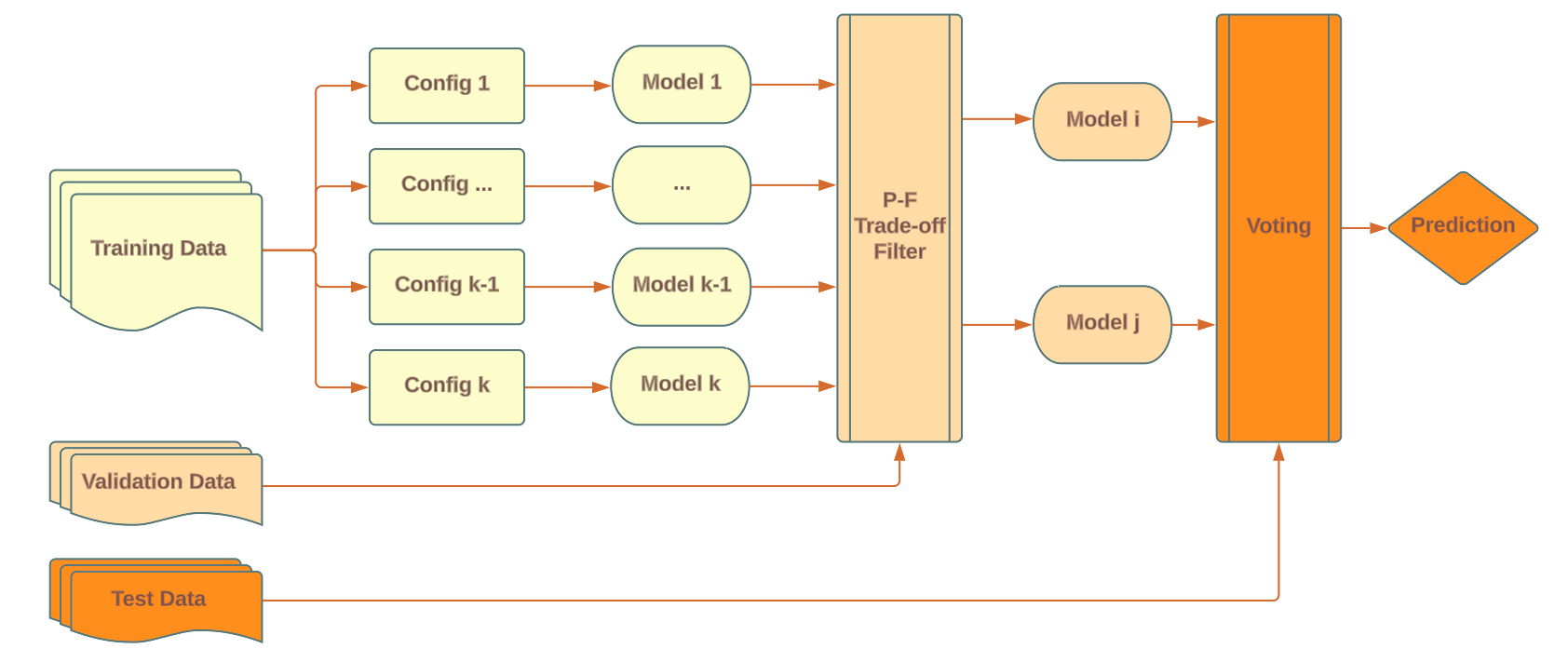}
\caption{
The general flowchart of the GroupCART Algo.  } 
\label{fig:framework}
\end{figure*}

Many prior research studies are searching for means to incorporate fairness constraints in decision tree algorithms. Kamiran et al. \cite{kamiran2010discrimination} in 2010 first proposed to utilize both the IG on the sensitive/protected attribute (namely IGS) and the IG on the class attribute (namely IGC) as the splitting criteria. Providing three alternatives of combining IGC and IGS, the algorithm aims to guide decision trees to split on the attribute that is homogeneous w.r.t. class attribute but heterogeneous w.r.t. sensitive attribute. In 2019, Zhang et al. extended this definition to "fairness information gain" in their paper, which proposed an adaptive fairness-aware decision tree Algorithm.
Similarly, Ranzato et al. \cite{ranzato2021fairness} proposed a tree-based framework that focuses on incorporating individual fairness constraints in terms of distance to the decision boundary. The approach proposed in their paper iteratively refines the decision tree via a predefined abstract domain that represents the fairness criteria. The decision tree uses abstract transformers to enforce fairness during the tree growth process.

\subsection{Ensemble Learning}\label{el}
Ensemble learning has emerged as a widely used technique in the field of machine learning. It involves combining multiple models/methods to form a single, stronger predictive pipeline. In the field of machine learning fairness, prior researchers also explored the applicability of ensemble learning. Nina et al.~\cite{grgic2017fairness} first proposed a solution in which it is believed that algorithmic discrimination could be mitigated via bagging a group of baseline models. Although their proposal remains at the theoretical stage, other researchers were still inspired and continue to propose practical solutions using ensemble learning techniques. Dheeraj et al. \cite{bhaskaruni2019improving}, for example, proposed using situational testing based on k-NN to identify unfair predictions, which will be later corrected through their ensemble strategy. Chakraborty et al. \cite{chakraborty2019software}, on the other hand, proposed an ensemble of pre-processing and in-processing methods to identify and mitigate potential bias within training data. More recently, Chen et al. proposed MAAT, an ensemble learning framework that aims to achieve superior performance-fairness trade-off by training and aggregating different models, respectively, oriented by predictive performance and fairness constraints.

\subsection{Multi-objective Optimization}\label{moo}
Multi-objective optimization has emerged as a valuable technique in software engineering, enabling the simultaneous optimization of multiple conflicting objectives during the software development process. 
Traditionally, software engineering tasks have been focused on single-objective optimization, with the aim of optimizing a specific metric or goal. However, in complex software systems, multiple objectives often need to be considered simultaneously, such as maximizing performance while minimizing resource utilization or enhancing reliability alongside minimizing development cost. Multi-objective optimization provides a systematic approach to exploring the trade-offs among these conflicting objectives, allowing software engineers to identify and select solutions that represent a well-balanced compromise.


Fairness-aware multi-objective optimization involves exploring the trade-offs between accuracy, fairness, and potentially other objectives, such as interpretability or efficiency. 
This optimization process aims to find a set of model configurations that provide a well-balanced compromise between these objectives. While there always exists an optimal solution in a single-objective optimization task (where a better solution can be determined by a single criterion), there might not be any optimal solution in a multi-objective optimization task. This might be the situation where no configuration is best for all objectives.
Therefore, we need to quantify and evaluate the overall quality of a configuration in a different way, called the {\em domination} relationship. There are two main types of domination relationship: binary domination~\cite{holland1992genetic}, and continuous domination~\cite{laumanns2002combining}. According to the definition of binary domination, a configuration $c_1$ is binary dominant over $c_2$ iff:
\begin{equation}\label{b-dom}
\begin{gathered}
    y_{1,i} \leq y_{2,i} \forall i \in \{ 1, 2, ..., n \}     \text{ and} \\
    y_{1,j} < y_{2,j} \text{ for, at least, one } j \in  \{ 1, 2, ..., n \}     
\end{gathered}
\end{equation}
In contrast, continuous domination defines that a configuration $c_1$ is dominant over $c_2$ iff:
\begin{equation}\label{c-dom}
loss(c_1, c_2) < loss(c_2, c_1)
\end{equation}
where the loss function is defined as:
\begin{equation}\label{loss}
\begin{gathered}
loss(c_1, c_2) = \Sigma_{j=1}^{n} -e^{(y_{2,j}-y_{1,j}) }\times \frac{1}{n}\\
\text{where } y_{i,j} \text{ is min-max normalized}
\end{gathered}
\end{equation}

Note that in both equations above, the default setting is that the lower performance measure $y$ is preferred.
By definition, a best (optimal) configuration is one that is not dominated by any other configuration, denoted as a non-dominated solution. The set containing all non-dominated solutions is called the Pareto frontier set. In short, the goal of a multi-objective optimization problem is to find as many optimal (non-dominated) solutions as possible while minimizing the evaluation cost (using fewer measurements).

\begin{algorithm}[t!]
\caption{Non-dominated Sorting (NDSorting) \label{algo2}}
\KwData{$C$ contains configurations awaiting to be sorted;
performance function $f$ maps configurations to the corresponding objective values.
}
\KwResult{A set of non-dominated configurations $C_{nd}$, also referred to as the Pareto frontier.}
\Begin{
    $C_{first} \gets$ $C$\\
    $C_{nd} \gets$ $C$\\
    \While{$C_{first} \neq \emptyset$}{
        $c_1 \gets C_{first}$.pop()\\
        $C_{second} \gets$ $C_{first}$\\
        \While{$C_{second} \neq \emptyset$}{
            $c_2 \gets C_{second}$.pop()\\
            // using either domination as defined in Eq. \ref{b-dom} or Eq. \ref{c-dom} \\
            \If{$f(c_1)$ dominates $f(c_2)$}{
                $C_{nd}$.remove($c_2$)\\
            }
            \If{$f(c_2)$ dominates $f(c_1)$}{
                $C_{nd}$.remove($c_1$)\\
            }
        }
    }
    return $C_{nd}$
}
\end{algorithm}

To compute the non-dominated solution set, the
{\bf non-dominated sorting} process is required. As described in Algo. \ref{algo2}, the algorithm iteratively compared every pair of candidates (in this case, configurations) to determine the domination relationship between each pair. The process has a runtime complexity of $O(MN^3)$ where $M$ is the number of objectives and $N$ is the population size. It should also be noted that NSGA-II~\cite{deb2002fast} proposed a fast and elitism sorting approach that reduced this complexity to $O(MN^2)$.  As described later, this paper follows the same sorting process used in NSGA-II.

\begin{algorithm}[t!]
\caption{GroupCART \label{algo3}}
\KwData{$N$ is the number of ensemble learners within GroupCART;
Objective function $f$ containing both fairness and performance metrics returns unary objective scores;
Training data $D_{train}$, validation data $D_{valid}$, and test data $D_{test}$.
}
\KwResult{A set of non-dominated models $M_{nd}$, also referred to as the Pareto frontier; Final predictions of the test data $Y_{pred}$.}
\Begin{
    \For{$i \gets 1$ \textbf{to} $N$}{
        $M_{i} \gets$ Model.initialize()\\
        // Configure the model with a distinct ratio between the weight of IGS and that of IGC. \\
        $M_{i}.IGS.weights \gets$ $\frac{i}{N}$\\
        $M_{i}.IGC.weights \gets$ $1 - \frac{i}{N}$\\
        $M_{i}$.fit($D_{train}$)\\
        $M_{candidates}$.append($ M_{i}$.evaluate($D_{valid}$))\\
    }
    $M_{nd} \gets NDSorting(M_{candidates},f)$\\
    $Y_{pred} \gets M_{nd}$.MajorityVoting($D_{test}$)\\
    return $M_{nd}, Y_{pred}$
}
\end{algorithm}

\begin{algorithm}[t!]
\caption{GroupCART\_AdaBoost \label{algo4}}
\KwData{$N$ is the number of ensemble learners within GroupCART; Objective function $f$ containing both fairness and performance metrics returns unary objective scores;
Training data $D_{train}$, validation data $D_{valid}$, and test data $D_{test}$.
}
\KwResult{A set of non-dominated models $M_{nd}$, also referred to as the Pareto frontier; Final predictions of the test data $Y_{pred}$.}
\Begin{
    Initialize instance weights $w_0$ for $D_{train}$ uniformly\\
    \For{$i \gets 1$ \textbf{to} $N$}{
        $M_{i} \gets$ Model.initialize()\\
        // Configure the model with a distinct ratio between the weight of IGS and that of IGC. \\
        $M_{i}.IGS.weights \gets$ $\frac{i}{N}$\\
        $M_{i}.IGC.weights \gets$ $1 - \frac{i}{N}$\\
        
        // Train model with weighted dataset\\
        $M_{i}$.fit($D_{train}, w_{i-1}$)\\
        
        // Calculate error rate $\epsilon_i$ for $M_{i}$ on $D_{train}$ based on $w_{i-1}$ \\
        $\epsilon_i \gets \sum_{x \in D_{train}} w_{i-1}(x) \cdot \mathbf{1}[M_{i}(x) \neq y]$\\
        
        // Calculate model weight  \\
        $\alpha_i \gets \frac{1}{2} \ln\left(\frac{1 - \epsilon_i}{\epsilon_i + \epsilon}\right),\quad \text{where } \epsilon \approx 10^{-10}$\\
        
        // Update instance weights $w_i$ for next iteration \\
        \ForEach{$x \in D_{train}$}{
            $w_i(x) \gets w_{i-1}(x) \cdot \exp(\alpha_i \cdot \mathbf{1}[M_{i}(x) \neq y])$\\
        }
        Normalize $w_i$ so that $\sum w_i = 1$\\
        
        $M_{candidates}$.append($ M_{i}$.evaluate($D_{valid}$))\\
    }
    $M_{nd} \gets NDSorting(M_{candidates},f)$\\
    $Y_{pred} \gets M_{nd}$.WeightedVoting($D_{test}, \{\alpha_i\})$\\
    return $M_{nd}, Y_{pred}$
}
\end{algorithm}

\section{Our Approach: GroupCART}
We propose a multi-level framework that first narrows down the pool of candidate models for the performance-fairness trade-off and then moderates the predictions via majority voting. The visualized flow chart for our approach is shown in Figure~\ref{fig:framework}, and the algorithmic description is shown in Algo. \ref{algo3} and Algo. \ref{algo4} (with AdaBoost).
This section describes, in parts, the design of our approach. 

In order to adhere to current empirical standards in SE,
that description adheres to the concepts recommended in the ACM standards documents on empirical  SE for optimization studies\footnote{\url{https://github.com/acmsigsoft/EmpiricalStandards/blob/master/docs/OptimizationStudies.md}}.

\subsection{Fairness as Hyper-parameter Configuration}
Prior studies illustrate how hyperparameters influence models' behavior in the performance-fairness trade-off \cite{cruz2021promoting,tizpaz2022fairness,hort2024search}. As shown in Figure. \ref{pareto}, different configuration options can result in enormous states of models on the spectrum of performance versus fairness.
Thus, model configurations that bring optimal performance-fairness trade-offs are in demand. However, 
constructing and assessing candidate configurations can be time-consuming given the number of options available for tuning. Therefore, in GroupCART, we propose an intuitive and straightforward way for the construction of configurations. That is, we construct a family of fairness-aware decision trees (FDTs) by tuning the internal weight ratio between the IGC and IGS (as elaborated in \S\ref{fdt}) when determining a split inside the tree.
By controlling the threshold $\theta$, we can build models that specifically focus on certain selected fairness metrics (accuracy, f1, aod, eod, etc.). Theoretically, this process provides us with models of different extend of fairness. 

\subsection{Multi-objective Configuration Optimization}
After constructing candidate models of different performance-fairness configurations, GroupCART filters out a subset of models that achieve superior trade-offs on the validation data. This subset is called the Pareto frontier in the field of multi-objective optimization. In the experiment conducted in this paper, we use a mixture of four performance/fairness scores to assess the domination relationship among candidate solutions: accuracy, f1 score, average odds difference (AOD), and disparate impact (DI). 

An observation from the literature on multi-objective optimization is that the Pareto frontier solutions are usually not unique due to the definition of binary domination, especially with increasing numbers of objectives \cite{zitzler1999multiobjective,zitzler2001spea2,deb2002fast,nair2018finding}. In our scenario, this means that if multiple fairness/performance metrics are included in the optimization process, it becomes more likely to result in many candidate models, as seen in Figure. \ref{fig:framework}. Therefore, we choose to resolve such confusion by adapting continuous domination when calculating the Pareto frontier. As described in Eq. \ref{c-dom}, continuous domination provides a unified value representing the dominance status of a candidate solution on the objective space. 

Moreover, as this paper attempts to tackle the performance-fairness trade-offs as a software configuration optimization task, we would also like to measure the quality of the resulting models via certain unified metrics used in multi-objective optimization. Therefore, we calculated the {\it d2h}  \cite{agrawal2019dodge} (distance to heaven) scores using all the performance and fairness scores introduced above. The equation assumes a ``heaven“ model that achieves optimal scores in all metrics and computes the Euclidean distance between the existing model and the ``heaven" model. A lower score {\it d2h} indicates that a model is closer to the ideal optimum. 
As shown below, $Optimum$ contains the ideal objective scores for each metric (e.g., the optimum for accuracy is 100).
\begin{equation}\label{d2h}
\begin{gathered}
d2h = \sqrt{\Sigma_{j=1}^{n} (Optimum_{j}-metric_{j})^2}\\
\text{where } m_{j} \text{ is the score for the $j_{th}$ performance/fairness metric.}
\end{gathered}
\end{equation}
The equation simply measures the $n$-dimensional Euclidean distance between the ideal scores and actual objective scores of a model, where $n$ is the number of metrics (either performance or fairness) included in the evaluation.

\subsection{Ensemble Voting}
With the use of continuous domination in the multi-objective optimization process, it still remains possible that multiple configurations/solutions may exist in the Pareto frontier set. Therefore, the last stage of the GroupCART framework is to generate final predictions through majority voting on predictions from various models constructed using the Pareto frontier configurations. 

As later observed in the experiment, ensemble voting is rather helpful as GroupCART usually finds more than one Pareto frontier configuration. 
Compared to traditional hyper-parameter optimization methods, including ones from Cruz's work, GroupCART constructs candidate pools of far smaller sizes, which implies better computational complexity. This is because the initial candidate configurations in GroupCART are stratified based on fairness-aware criteria, which is the information gain on protected attributes in this case.

\section{Experiment Setup}\label{experiment}

In this section, we describe the data preparation for the experiment as well as the general setup of the experiment.

In order to adhere to current empirical standards in SE,
that description adheres to the experimental design principles recommended in the ACM standards documents on empirical SE for optimization studies\footnote{\url{https://github.com/acmsigsoft/EmpiricalStandards/blob/master/docs/OptimizationStudies.md}}.

\subsection{Data}\label{data}
This paper uses data sets collected that are widely used in prior related research (see Table~\ref{tab:dataset}). After data collection, we first need to pre-process the data. For most of the datasets used in this paper (German, Bank, Heart, etc.), no feature engineering is required because either the features are all numerical or a standard procedure is adopted by all prior practitioners. As for others, in this experiment, a standardized pre-processing procedure is adopted, following guidelines from the AIF360 repository~\cite{bellamy2018ai}.
Finally, we apply min-max scaling (scale numerical values in the range of $[0,1]$ by the minimum and maximum values in each feature) to transform each data set. For each experiment trial, we divide the data into 70\% training data, 10\% validation data, and 20\% testing data, using the same set of random seeds in all methods to control the comparison variable. We repeat this procedure 20 times for statistical analysis.

\begin{table}[t]
\centering
\caption{Performance metrics based on binary confusion matrix.}
\begin{tabular}{cc}
\toprule
Metrics   & Definition                                    \\ \hline
Accuracy   & (TP+TN) / (TP+TN+FP+FN)                         \\ 
Precision & TP / (TP+FP)                                    \\ 
Recall    & TP / (TP+FN)                                    \\ \
F1 score  & 2 $\times$ (Precision $\times$ Recall)$/$(Precision + Recall) \\ 
\bottomrule
\end{tabular}
\label{tab:metrics2}
\end{table}

\subsection{Baselines}
In our experiment, we have four other baseline methods:
\begin{itemize}
    \item {\bf CART}: Our proposed approach can be considered as a specifically modified decision tree, where the information gain of class labels is not the sore splitting criterion. Thus, we would like to compare our approach to the original, normal version of CART.
    \item {\bf EGR}: As introduced previously in \S\ref{bias-mitigation}, Exponentiated gradient reduction (EGR) is an in-processing technique that reduces fair classification to a sequence of cost-sensitive classification problems \cite{agarwal2018reductions}.
    \item {\bf Fax-AI}: Fax-AI \cite{grabowicz2022marrying} is a post-processing method that mitigates model bias by eliminating potential proxy discrimination. Fax-AI limits the usage of certain features, as those features are believed to have a greater likelihood of serving as surrogates for protected attributes.
    \item {\bf MAAT}: MAAT \cite{chen2022maat} is one of the most recent ensemble-based fairness learning methods. As shown in its paper, MAAT has outperformed several previous SOTA methods, including Fairway and Fair-SMOTE, by obtaining a better performance-fairness trade-off in many studied datasets.
    \item {\bf FairMask}: FairMask \cite{peng2022fairmask} is a recent bias mitigation method that generates synthesized protected attributes (PAs) parallel to the model training phase, and synthetic attributes are used to replace real protected attributes during inference time. FairMask is designed to achieve not only distributive fairness but also perfect individual fairness.
\end{itemize}
We use the same data splitting style as was performed on {\IT} for the methods where validation data is needed. As for others, the data is only divided into training and testing sets, and we ensure that in each trial, the same testing set is used by {\IT} and other baselines.

\subsection{Experimental Rig}\label{rig}
One summary of this paper is that previous work exploring ensemble learning and fairness used
too few ensembles. More specifically, we will show that the ensembles of size two used by MAAT can be improved by using more members in the ensembles.
This raises the question ``If MAAT's ensembles are too small, when are ensembles big enough?''. 
Some of our runs can be slow (since we are currently working with prototypes not applicable for distributed computation), so when exploring ``big'', we first reflected on early stopping rules that would let us explore bigger ensembles, but not cripple our rig with excess processing.

One such early stopping rule is to use the Flip Rate from \S\ref{rw}. Recall that the Flip Rate (FR) computes the ratio of the population whose prediction outcomes are flipped when reversing their protected attributes.  Models that are perfectly fair are expected to obtain a FLIP rate of zero.  Hence, in this paper, we keep doubling the size of the ensembles, stopping after the point where
FR $=0$. 
In practice, this means that we will generate ensembles of sizes 5,10, and 20 (as the FR decreases to 0 at the size of 20).

\subsection{Evaluation Criteria}
The overall effectiveness of a bias mitigation method can be assessed from two aspects: predictive performance and fairness measurements.
To evaluate the predictive performance in a binary classification task, we use metrics computed by the confusion matrix: accuracy, precision, recall, and F1 score. These criteria
are selected since (a) they are widely used in both software analytics~\cite{local_tse,menzies2014sharing} and fairness literature~\cite{9286091,Biswas_2020,hardt2016equality,pleiss2017fairness,zhang2020white}
and (b) the latest benchmark methods in our experiment, such as MAAT \cite{chen2022maat} and FairMask \cite{peng2022fairmask}, use the exact same set of criteria.
                The definitions of the performance metrics are shown in Table~\ref{tab:metrics2}.

Meanwhile, to quantify the effectiveness of mitigating bias, we use the fairness metrics introduced in Table~\ref{tab:metrics}. Group fairness metrics are designed to evaluate whether different social groups, as identified by their protected attributes, receive statistically similar prediction results by the classification model. The individual fairness metric (flip rate in this case) on the other hand is designed based on the intuition of procedural justice. By definition, when individuals are similar to each other regardless of the protected attributes, they shall receive similar prediction outcomes (in this case of binary classification, the same outcome). To assess this criterion, we use the following situation-test tactic:
\begin{itemize}
    \item For each instance in testing data, flip the protected attribute.
    \item Pass the edited data instances into the classification model
    \item Record the times when the new prediction outcome differs from the original one.
\end{itemize}
While other fairness metrics assess distributive fairness from a sampled population, this metric is specifically designed to measure individual fairness as used in prior research \cite {chakraborty2019software,peng2022fairmask}.

\subsection{Statistical Analysis}\label{stats}

In order to tell whether our proposed approach is statistically significantly better than other baselines, we use a non-parametric significance test along with a non-parametric effect size test. 
In this experiment, we use the Scott-Knott test~\cite{mittas2012ranking} that sorts the list of treatments by their median scores. After sorting, the list is split into two sub-lists. The objective for such a split is to maximize the expected value of differences $E(\Delta)$ in the observed performances before and after division~\cite{xia2018hyperparameter}:
\begin{equation}
    E(\Delta) = \frac{|l_1|}{|l|}abs(E({l_1}) - E({l}))^2 + \frac{|l_2|}{|l|}abs(E({l_2}) - E({l}))^2
\end{equation}
where $|l_1|$ means the size of list $l_1$.

The Scott-Knott test assigns ranks to each result set; the higher the rank, the better the result. Two results will be ranked the same if the difference between the distributions is not significant.
In this expression, Cliff's Delta estimates the probability that a value in the list $A$ is greater than a value in the list $B$, minus the reverse probability~\cite{macbeth2011cliff}. A division passes this hypothesis test if it is not a ``small'' effect ($Delta \geq 0.147$). 
This hypothesis test and its effect sizes are supported by Hess and Kromery~\cite{hess2004robust}.

\begin{table}[t!]
\centering
\caption{Results for RQ1. The table shows, in the descending order of delta, the differences in {\it d2h} scores obtained by {\IT} using the default weight versus using the optimal weight combination.}
\begin{tabular}{rrrrl}
   & \multicolumn{2}{c}{Distance to heaven ({\it d2h})}\\\cline{2-3}
                
              & Default weight & Optimized weight & Delta \%  &\\ \midrule
Bank: Age    & .77             & .51             & 51.0      &  \\
Adult: Sex   & 1.14            & .87             & 31.0      &  \\
German: Sex  & .60             & .46             & 30.4     &  \\  
Adult:Race   & .84             & .75             & 12.0     &   \\
\rowcolor{blue!10}Heart: Age   & .77             & .70      &         10.0 & median delta      \\
MEPS: Race   & 1.10            & 1.01            & 8.9      &   \\
Compas: Sex  & .69             & .65             & 6.2      &   \\
Default: Sex & 1.01            & .96             & 5.2      &   \\
Compas:Race  & .69             & .69             & 0.0      &   \\\hline
Average      & .85 & .73   \\
\end{tabular}\label{tab:rq1}
\end{table}

\subsection{Research Questions}
We assess the merits of {\IT} centering around the following research questions:

\begin{itemize}
    \item {\bf RQ1:} {\it How does multi-objective optimization affect the effectiveness of GroupCART?}
    Instead of applying a fixed weight ratio between IGS and IGC, our approach generates a set of configurations, each representing a model with a different IGS/IGC weight ratio. In this RQ, we evaluate the effectiveness of this tactic by comparing the default choice of weight with the optimal weight combination returned by our multi-objective optimization.
    \item {\bf RQ2:} {\it How does the scale of ensemble learners affect the effectiveness of GroupCART?}
    How large should our ensemble learner be in order to achieve a better trade-off between performance and fairness? In this RQ, we examine how the size of ensemble learning affects our approach and demonstrate preliminary experimental results. 
    \item {\bf RQ3:} {\it How effective is GroupCART compared to state-of-the-art methods?} 
    In this RQ, we compare GroupCART with four baselines as introduced in the previous section. We analyze whether our proposed approach can obtain comparable or superior results.
    \item {\bf RQ4:} {\it How effective is GroupCART when dealing with multiple protected attributes?} Some data sets have more than one protected attribute. In real-life practice, it would be preferred if the software model could handle multiple protected attributes simultaneously. In this RQ, we evaluate the effectiveness of our approach in scenarios with multiple protected attributes.
\end{itemize}

\section{Results}
To assess the effectiveness of our proposed approach compared to other benchmark methods, we design the evaluation of the experiment around four research questions (RQs).

\begin{blockquote}
\textbf{RQ1}: How does multi-objective optimization affect the effectiveness of GroupCART?
\end{blockquote}
To evaluate models in performance-fairness optimization, we use the {\it d2h} (distance to heaven) scores to indicate the general effectiveness of a model.
Table~\ref{tab:rq1} contains the {\it d2h} scores achieved in each of the empirical datasets when using (a) the default hyper-parameter configuration, and (b) the optimized configurations returned by GroupCART. We show that, in only one out of nine cases, the optimal configuration is the same as the default (see the last row of 
Table~\ref{tab:rq1}
In other cases, the default configuration is significantly worse (by up to 50\%) than the optimal ones. 

Thus, our answer to RQ1 is: {\bf In most cases, multi-objective optimization is required since
the off-the-shelf default parameters required optimizing to achieve better trade-offs (with lower {\it d2h}).}

\begin{table*}[!htbp]
\small
\centering
\caption{Results for RQ3.  For all performance metrics, greater is better; for all fairness metrics, smaller is better. We repeat the experiment for 20 runs for each dataset and report the median values in percentage. The ranks indicated by colors are determined by the Scott-Knott test.
}
\resizebox{\textwidth}{!}{%
\begin{tabular}{|c|c|cccc|ccccc|}
\hline
Dataset &
  Method &
  Accuracy &
  Precision &
  Recall &
  F1 &
  AOD &
  EOD &
  SPD &
  DI &
  FR \\ \hline
 &
  CART &
  \cellcolor[HTML]{9B9B9B}84 &
  \cellcolor[HTML]{9B9B9B}72 &
  \cellcolor[HTML]{EFEFEF}51 &
  \cellcolor[HTML]{C0C0C0}60 &
  \cellcolor[HTML]{EFEFEF}13 &
  \cellcolor[HTML]{EFEFEF}36 &
  \cellcolor[HTML]{EFEFEF}22 &
  90 &
  \cellcolor[HTML]{EFEFEF}17 \\
 &
  EGR &
  \cellcolor[HTML]{9B9B9B}84 &
  \cellcolor[HTML]{9B9B9B}76 &
  \cellcolor[HTML]{EFEFEF}53 &
  \cellcolor[HTML]{C0C0C0}63 &
  \cellcolor[HTML]{9B9B9B}2 &
  \cellcolor[HTML]{9B9B9B}3 &
  \cellcolor[HTML]{9B9B9B}11 &
  \cellcolor[HTML]{C0C0C0}52 &
  \cellcolor[HTML]{C0C0C0}3 \\
 &
  Fax-AI &
  \cellcolor[HTML]{C0C0C0}81 &
  \cellcolor[HTML]{EFEFEF}61 &
  \cellcolor[HTML]{C0C0C0}61 &
  \cellcolor[HTML]{C0C0C0}61 &
  \cellcolor[HTML]{9B9B9B}1 &
  \cellcolor[HTML]{C0C0C0}6 &
  \cellcolor[HTML]{C0C0C0}17 &
  \cellcolor[HTML]{C0C0C0}54 &
  \cellcolor[HTML]{9B9B9B}0 \\
 &
  MAAT &
  \cellcolor[HTML]{C0C0C0}80 &
  \cellcolor[HTML]{9B9B9B}74 &
  \cellcolor[HTML]{9B9B9B}76 &
  \cellcolor[HTML]{9B9B9B}75 &
  \cellcolor[HTML]{C0C0C0}6 &
  \cellcolor[HTML]{9B9B9B}7 &
  \cellcolor[HTML]{C0C0C0}16 &
  \cellcolor[HTML]{EFEFEF}68 &
  \cellcolor[HTML]{C0C0C0}7 \\
 &
  FairMask &
  \cellcolor[HTML]{C0C0C0}81 &
  \cellcolor[HTML]{EFEFEF}63 &
  \cellcolor[HTML]{C0C0C0}60 &
  \cellcolor[HTML]{C0C0C0}62 &
  \cellcolor[HTML]{9B9B9B}2 &
  \cellcolor[HTML]{9B9B9B}4 &
  \cellcolor[HTML]{C0C0C0}17 &
  \cellcolor[HTML]{EFEFEF}58 &
  \cellcolor[HTML]{9B9B9B}0 \\
\multirow{-6}{*}{Adult: Sex} &
  GroupCART &
  \cellcolor[HTML]{9B9B9B}83 &
  \cellcolor[HTML]{C0C0C0}66 &
  \cellcolor[HTML]{EFEFEF}55 &
  \cellcolor[HTML]{C0C0C0}60 &
  \cellcolor[HTML]{9B9B9B}2 &
  \cellcolor[HTML]{9B9B9B}5 &
  \cellcolor[HTML]{9B9B9B}9 &
  \cellcolor[HTML]{9B9B9B}47 &
  \cellcolor[HTML]{9B9B9B}0 \\ \hline
 &
  CART &
  \cellcolor[HTML]{9B9B9B}84 &
  \cellcolor[HTML]{9B9B9B}72 &
  \cellcolor[HTML]{EFEFEF}51 &
  \cellcolor[HTML]{C0C0C0}60 &
  \cellcolor[HTML]{C0C0C0}3 &
  \cellcolor[HTML]{EFEFEF}9 &
  \cellcolor[HTML]{C0C0C0}9 &
  \cellcolor[HTML]{EFEFEF}47 &
  \cellcolor[HTML]{C0C0C0}6 \\
 &
  EGR &
  \cellcolor[HTML]{C0C0C0}81 &
  \cellcolor[HTML]{C0C0C0}63 &
  \cellcolor[HTML]{C0C0C0}62 &
  \cellcolor[HTML]{C0C0C0}62 &
  \cellcolor[HTML]{C0C0C0}2 &
  \cellcolor[HTML]{9B9B9B}2 &
  \cellcolor[HTML]{C0C0C0}9 &
  \cellcolor[HTML]{C0C0C0}34 &
  \cellcolor[HTML]{9B9B9B}0 \\
 &
  Fax-AI &
  \cellcolor[HTML]{C0C0C0}80 &
  \cellcolor[HTML]{EFEFEF}60 &
  \cellcolor[HTML]{EFEFEF}55 &
  \cellcolor[HTML]{C0C0C0}58 &
  \cellcolor[HTML]{C0C0C0}3 &
  \cellcolor[HTML]{9B9B9B}1 &
  \cellcolor[HTML]{9B9B9B}6 &
  \cellcolor[HTML]{9B9B9B}26 &
  \cellcolor[HTML]{9B9B9B}0 \\
 &
  MAAT &
  \cellcolor[HTML]{C0C0C0}80 &
  \cellcolor[HTML]{9B9B9B}74 &
  \cellcolor[HTML]{9B9B9B}76 &
  \cellcolor[HTML]{9B9B9B}75 &
  \cellcolor[HTML]{C0C0C0}3 &
  \cellcolor[HTML]{C0C0C0}3 &
  \cellcolor[HTML]{9B9B9B}6 &
  \cellcolor[HTML]{C0C0C0}35 &
  \cellcolor[HTML]{C0C0C0}10 \\
 &
  FairMask &
  \cellcolor[HTML]{C0C0C0}81 &
  \cellcolor[HTML]{C0C0C0}63 &
  \cellcolor[HTML]{C0C0C0}61 &
  \cellcolor[HTML]{C0C0C0}62 &
  \cellcolor[HTML]{C0C0C0}2 &
  \cellcolor[HTML]{9B9B9B}2 &
  \cellcolor[HTML]{C0C0C0}8 &
  \cellcolor[HTML]{C0C0C0}33 &
  \cellcolor[HTML]{9B9B9B}0 \\
\multirow{-6}{*}{Adult: Race} &
  GroupCART &
  \cellcolor[HTML]{9B9B9B}83 &
  \cellcolor[HTML]{C0C0C0}65 &
  \cellcolor[HTML]{C0C0C0}59 &
  \cellcolor[HTML]{C0C0C0}60 &
  \cellcolor[HTML]{9B9B9B}1 &
  \cellcolor[HTML]{C0C0C0}4 &
  \cellcolor[HTML]{9B9B9B}6 &
  \cellcolor[HTML]{C0C0C0}39 &
  \cellcolor[HTML]{9B9B9B}0 \\ \hline
 &
  CART &
  \cellcolor[HTML]{9B9B9B}65 &
  \cellcolor[HTML]{9B9B9B}67 &
  \cellcolor[HTML]{C0C0C0}69 &
  \cellcolor[HTML]{C0C0C0}68 &
  \cellcolor[HTML]{C0C0C0}4 &
  \cellcolor[HTML]{C0C0C0}6 &
  \cellcolor[HTML]{EFEFEF}11 &
  \cellcolor[HTML]{EFEFEF}17 &
  \cellcolor[HTML]{EFEFEF}22 \\
 &
  EGR &
  \cellcolor[HTML]{9B9B9B}66 &
  \cellcolor[HTML]{9B9B9B}66 &
  \cellcolor[HTML]{9B9B9B}78 &
  \cellcolor[HTML]{9B9B9B}71 &
  \cellcolor[HTML]{9B9B9B}2 &
  \cellcolor[HTML]{9B9B9B}3 &
  \cellcolor[HTML]{9B9B9B}5 &
  \cellcolor[HTML]{9B9B9B}8 &
  \cellcolor[HTML]{C0C0C0}11 \\
 &
  Fax-AI &
  \cellcolor[HTML]{C0C0C0}61 &
  \cellcolor[HTML]{C0C0C0}63 &
  \cellcolor[HTML]{C0C0C0}67 &
  \cellcolor[HTML]{EFEFEF}65 &
  \cellcolor[HTML]{C0C0C0}4 &
  \cellcolor[HTML]{C0C0C0}4 &
  \cellcolor[HTML]{C0C0C0}7 &
  \cellcolor[HTML]{C0C0C0}12 &
  \cellcolor[HTML]{9B9B9B}0 \\
 &
  MAAT &
  \cellcolor[HTML]{9B9B9B}67 &
  \cellcolor[HTML]{9B9B9B}66 &
  \cellcolor[HTML]{C0C0C0}66 &
  \cellcolor[HTML]{EFEFEF}66 &
  \cellcolor[HTML]{EFEFEF}7 &
  \cellcolor[HTML]{C0C0C0}5 &
  \cellcolor[HTML]{C0C0C0}9 &
  \cellcolor[HTML]{C0C0C0}12 &
  \cellcolor[HTML]{C0C0C0}19 \\
 &
  FairMask &
  \cellcolor[HTML]{C0C0C0}61 &
  \cellcolor[HTML]{C0C0C0}63 &
  \cellcolor[HTML]{C0C0C0}65 &
  \cellcolor[HTML]{EFEFEF}65 &
  \cellcolor[HTML]{C0C0C0}4 &
  \cellcolor[HTML]{C0C0C0}5 &
  \cellcolor[HTML]{9B9B9B}5 &
  \cellcolor[HTML]{9B9B9B}9 &
  \cellcolor[HTML]{9B9B9B}0 \\
\multirow{-6}{*}{Compas: Sex} &
  GroupCART &
  \cellcolor[HTML]{9B9B9B}65 &
  \cellcolor[HTML]{9B9B9B}66 &
  \cellcolor[HTML]{9B9B9B}74 &
  \cellcolor[HTML]{9B9B9B}70 &
  \cellcolor[HTML]{9B9B9B}2 &
  \cellcolor[HTML]{9B9B9B}3 &
  \cellcolor[HTML]{C0C0C0}8 &
  \cellcolor[HTML]{C0C0C0}11 &
  \cellcolor[HTML]{9B9B9B}0 \\ \hline
 &
  CART &
  \cellcolor[HTML]{9B9B9B}65 &
  \cellcolor[HTML]{9B9B9B}67 &
  \cellcolor[HTML]{C0C0C0}69 &
  \cellcolor[HTML]{9B9B9B}68 &
  \cellcolor[HTML]{C0C0C0}3 &
  \cellcolor[HTML]{C0C0C0}10 &
  \cellcolor[HTML]{C0C0C0}14 &
  \cellcolor[HTML]{EFEFEF}22 &
  \cellcolor[HTML]{EFEFEF}21 \\
 &
  EGR &
  \cellcolor[HTML]{9B9B9B}66 &
  \cellcolor[HTML]{9B9B9B}68 &
  \cellcolor[HTML]{C0C0C0}70 &
  \cellcolor[HTML]{9B9B9B}69 &
  \cellcolor[HTML]{9B9B9B}2 &
  \cellcolor[HTML]{9B9B9B}4 &
  \cellcolor[HTML]{9B9B9B}5 &
  \cellcolor[HTML]{9B9B9B}8 &
  \cellcolor[HTML]{C0C0C0}15 \\
 &
  Fax-AI &
  \cellcolor[HTML]{C0C0C0}61 &
  \cellcolor[HTML]{C0C0C0}63 &
  \cellcolor[HTML]{EFEFEF}67 &
  \cellcolor[HTML]{C0C0C0}65 &
  \cellcolor[HTML]{C0C0C0}3 &
  \cellcolor[HTML]{9B9B9B}6 &
  \cellcolor[HTML]{C0C0C0}10 &
  \cellcolor[HTML]{C0C0C0}16 &
  \cellcolor[HTML]{9B9B9B}0 \\
 &
  MAAT &
  \cellcolor[HTML]{9B9B9B}67 &
  \cellcolor[HTML]{9B9B9B}67 &
  \cellcolor[HTML]{EFEFEF}67 &
  \cellcolor[HTML]{C0C0C0}66 &
  \cellcolor[HTML]{EFEFEF}16 &
  \cellcolor[HTML]{C0C0C0}12 &
  \cellcolor[HTML]{EFEFEF}18 &
  \cellcolor[HTML]{EFEFEF}26 &
  \cellcolor[HTML]{EFEFEF}20 \\
 &
  FairMask &
  \cellcolor[HTML]{C0C0C0}61 &
  \cellcolor[HTML]{C0C0C0}63 &
  \cellcolor[HTML]{EFEFEF}65 &
  \cellcolor[HTML]{C0C0C0}65 &
  \cellcolor[HTML]{C0C0C0}3 &
  \cellcolor[HTML]{9B9B9B}6 &
  \cellcolor[HTML]{C0C0C0}10 &
  \cellcolor[HTML]{C0C0C0}16 &
  \cellcolor[HTML]{9B9B9B}0 \\
\multirow{-6}{*}{Compas: Race} &
  GroupCART &
  \cellcolor[HTML]{9B9B9B}65 &
  \cellcolor[HTML]{C0C0C0}65 &
  \cellcolor[HTML]{9B9B9B}74 &
  \cellcolor[HTML]{9B9B9B}69 &
  \cellcolor[HTML]{9B9B9B}2 &
  \cellcolor[HTML]{C0C0C0}10 &
  \cellcolor[HTML]{C0C0C0}13 &
  \cellcolor[HTML]{C0C0C0}18 &
  \cellcolor[HTML]{9B9B9B}0 \\ \hline
 &
  CART &
  \cellcolor[HTML]{C0C0C0}72 &
  \cellcolor[HTML]{C0C0C0}70 &
  \cellcolor[HTML]{EFEFEF}70 &
  \cellcolor[HTML]{C0C0C0}70 &
  \cellcolor[HTML]{C0C0C0}7 &
  \cellcolor[HTML]{EFEFEF}8 &
  \cellcolor[HTML]{EFEFEF}20 &
  43 &
  \cellcolor[HTML]{EFEFEF}12 \\
 &
  EGR &
  \cellcolor[HTML]{C0C0C0}69 &
  \cellcolor[HTML]{9B9B9B}77 &
  \cellcolor[HTML]{C0C0C0}78 &
  \cellcolor[HTML]{9B9B9B}78 &
  \cellcolor[HTML]{C0C0C0}5 &
  \cellcolor[HTML]{C0C0C0}7 &
  \cellcolor[HTML]{C0C0C0}9 &
  \cellcolor[HTML]{EFEFEF}12 &
  \cellcolor[HTML]{EFEFEF}10 \\
 &
  Fax-AI &
  \cellcolor[HTML]{EFEFEF}67 &
  \cellcolor[HTML]{9B9B9B}76 &
  \cellcolor[HTML]{C0C0C0}77 &
  \cellcolor[HTML]{9B9B9B}77 &
  \cellcolor[HTML]{C0C0C0}5 &
  \cellcolor[HTML]{C0C0C0}6 &
  \cellcolor[HTML]{C0C0C0}6 &
  \cellcolor[HTML]{C0C0C0}8 &
  \cellcolor[HTML]{9B9B9B}0 \\
 &
  MAAT &
  \cellcolor[HTML]{C0C0C0}71 &
  \cellcolor[HTML]{EFEFEF}60 &
  64 &
  \cellcolor[HTML]{EFEFEF}61 &
  \cellcolor[HTML]{C0C0C0}6 &
  \cellcolor[HTML]{C0C0C0}6 &
  \cellcolor[HTML]{C0C0C0}6 &
  \cellcolor[HTML]{9B9B9B}1 &
  \cellcolor[HTML]{C0C0C0}4 \\
 &
  FairMask &
  \cellcolor[HTML]{EFEFEF}67 &
  \cellcolor[HTML]{9B9B9B}77 &
  \cellcolor[HTML]{C0C0C0}76 &
  \cellcolor[HTML]{9B9B9B}76 &
  \cellcolor[HTML]{9B9B9B}3 &
  \cellcolor[HTML]{9B9B9B}4 &
  \cellcolor[HTML]{9B9B9B}4 &
  \cellcolor[HTML]{C0C0C0}6 &
  \cellcolor[HTML]{9B9B9B}0 \\
\multirow{-6}{*}{German: Sex} &
  GroupCART &
  \cellcolor[HTML]{9B9B9B}75 &
  \cellcolor[HTML]{C0C0C0}72 &
  \cellcolor[HTML]{9B9B9B}81 &
  \cellcolor[HTML]{9B9B9B}77 &
  \cellcolor[HTML]{C0C0C0}6 &
  \cellcolor[HTML]{C0C0C0}7 &
  \cellcolor[HTML]{C0C0C0}6 &
  \cellcolor[HTML]{EFEFEF}12 &
  \cellcolor[HTML]{9B9B9B}0 \\ \hline
 &
  CART &
  \cellcolor[HTML]{C0C0C0}88 &
  \cellcolor[HTML]{EFEFEF}52 &
  \cellcolor[HTML]{C0C0C0}53 &
  \cellcolor[HTML]{EFEFEF}54 &
  \cellcolor[HTML]{C0C0C0}5 &
  \cellcolor[HTML]{C0C0C0}7 &
  \cellcolor[HTML]{C0C0C0}12 &
  \cellcolor[HTML]{EFEFEF}89 &
  \cellcolor[HTML]{C0C0C0}2 \\
 &
  EGR &
  \cellcolor[HTML]{9B9B9B}91 &
  \cellcolor[HTML]{C0C0C0}65 &
  \cellcolor[HTML]{C0C0C0}53 &
  \cellcolor[HTML]{C0C0C0}58 &
  \cellcolor[HTML]{C0C0C0}4 &
  \cellcolor[HTML]{9B9B9B}6 &
  \cellcolor[HTML]{9B9B9B}7 &
  \cellcolor[HTML]{C0C0C0}68 &
  \cellcolor[HTML]{C0C0C0}3 \\
 &
  Fax-AI &
  \cellcolor[HTML]{C0C0C0}88 &
  \cellcolor[HTML]{EFEFEF}52 &
  \cellcolor[HTML]{C0C0C0}54 &
  \cellcolor[HTML]{EFEFEF}53 &
  \cellcolor[HTML]{9B9B9B}3 &
  \cellcolor[HTML]{C0C0C0}7 &
  \cellcolor[HTML]{9B9B9B}10 &
  \cellcolor[HTML]{EFEFEF}80 &
  \cellcolor[HTML]{9B9B9B}0 \\
 &
  MAAT &
  \cellcolor[HTML]{9B9B9B}90 &
  \cellcolor[HTML]{C0C0C0}77 &
  \cellcolor[HTML]{9B9B9B}78 &
  \cellcolor[HTML]{9B9B9B}77 &
  \cellcolor[HTML]{C0C0C0}4 &
  \cellcolor[HTML]{9B9B9B}5 &
  \cellcolor[HTML]{9B9B9B}9 &
  \cellcolor[HTML]{EFEFEF}79 &
  \cellcolor[HTML]{C0C0C0}4 \\
 &
  FairMask &
  \cellcolor[HTML]{C0C0C0}88 &
  \cellcolor[HTML]{EFEFEF}52 &
  \cellcolor[HTML]{C0C0C0}54 &
  \cellcolor[HTML]{EFEFEF}53 &
  \cellcolor[HTML]{9B9B9B}2 &
  \cellcolor[HTML]{9B9B9B}5 &
  \cellcolor[HTML]{9B9B9B}9 &
  \cellcolor[HTML]{C0C0C0}69 &
  \cellcolor[HTML]{9B9B9B}0 \\
\multirow{-6}{*}{Bank: Age} &
  GroupCART &
  \cellcolor[HTML]{EFEFEF}74 &
  \cellcolor[HTML]{9B9B9B}74 &
  \cellcolor[HTML]{9B9B9B}78 &
  \cellcolor[HTML]{9B9B9B}76 &
  \cellcolor[HTML]{EFEFEF}9 &
  \cellcolor[HTML]{9B9B9B}6 &
  \cellcolor[HTML]{EFEFEF}27 &
  \cellcolor[HTML]{9B9B9B}40 &
  \cellcolor[HTML]{9B9B9B}0 \\ \hline
 &
  CART &
  \cellcolor[HTML]{C0C0C0}80 &
  \cellcolor[HTML]{EFEFEF}41 &
  \cellcolor[HTML]{C0C0C0}44 &
  \cellcolor[HTML]{C0C0C0}42 &
  \cellcolor[HTML]{C0C0C0}4 &
  \cellcolor[HTML]{C0C0C0}8 &
  \cellcolor[HTML]{EFEFEF}12 &
  \cellcolor[HTML]{EFEFEF}46 &
  \cellcolor[HTML]{C0C0C0}3 \\
 &
  EGR &
  \cellcolor[HTML]{9B9B9B}85 &
  \cellcolor[HTML]{C0C0C0}57 &
  \cellcolor[HTML]{EFEFEF}35 &
  \cellcolor[HTML]{C0C0C0}43 &
  \cellcolor[HTML]{C0C0C0}4 &
  \cellcolor[HTML]{C0C0C0}9 &
  \cellcolor[HTML]{9B9B9B}4 &
  \cellcolor[HTML]{C0C0C0}29 &
  \cellcolor[HTML]{C0C0C0}4 \\
 &
  Fax-AI &
  \cellcolor[HTML]{C0C0C0}80 &
  \cellcolor[HTML]{EFEFEF}40 &
  \cellcolor[HTML]{C0C0C0}43 &
  \cellcolor[HTML]{C0C0C0}41 &
  \cellcolor[HTML]{C0C0C0}4 &
  \cellcolor[HTML]{EFEFEF}11 &
  \cellcolor[HTML]{EFEFEF}12 &
  \cellcolor[HTML]{EFEFEF}46 &
  \cellcolor[HTML]{9B9B9B}0 \\
 &
  MAAT &
  \cellcolor[HTML]{EFEFEF}77 &
  \cellcolor[HTML]{9B9B9B}64 &
  \cellcolor[HTML]{9B9B9B}69 &
  \cellcolor[HTML]{9B9B9B}65 &
  \cellcolor[HTML]{EFEFEF}8 &
  \cellcolor[HTML]{C0C0C0}9 &
  \cellcolor[HTML]{EFEFEF}12 &
  \cellcolor[HTML]{C0C0C0}35 &
  \cellcolor[HTML]{C0C0C0}4 \\
 &
  FairMask &
  \cellcolor[HTML]{C0C0C0}80 &
  \cellcolor[HTML]{EFEFEF}39 &
  \cellcolor[HTML]{C0C0C0}42 &
  \cellcolor[HTML]{C0C0C0}42 &
  \cellcolor[HTML]{9B9B9B}2 &
  \cellcolor[HTML]{C0C0C0}8 &
  \cellcolor[HTML]{C0C0C0}8 &
  \cellcolor[HTML]{C0C0C0}39 &
  \cellcolor[HTML]{9B9B9B}0 \\
\multirow{-6}{*}{Health: Age} &
  GroupCART &
  \cellcolor[HTML]{9B9B9B}83 &
  \cellcolor[HTML]{C0C0C0}49 &
  \cellcolor[HTML]{EFEFEF}38 &
  \cellcolor[HTML]{C0C0C0}43 &
  \cellcolor[HTML]{9B9B9B}2 &
  \cellcolor[HTML]{9B9B9B}6 &
  \cellcolor[HTML]{9B9B9B}2 &
  \cellcolor[HTML]{9B9B9B}12 &
  \cellcolor[HTML]{9B9B9B}1 \\ \hline
 &
  CART &
  \cellcolor[HTML]{9B9B9B}75 &
  \cellcolor[HTML]{9B9B9B}72 &
  \cellcolor[HTML]{C0C0C0}75 &
  \cellcolor[HTML]{C0C0C0}74 &
  \cellcolor[HTML]{C0C0C0}8 &
  \cellcolor[HTML]{EFEFEF}19 &
  \cellcolor[HTML]{EFEFEF}30 &
  \cellcolor[HTML]{EFEFEF}46 &
  \cellcolor[HTML]{C0C0C0}7 \\
 &
  EGR &
  \cellcolor[HTML]{C0C0C0}73 &
  \cellcolor[HTML]{C0C0C0}70 &
  \cellcolor[HTML]{C0C0C0}76 &
  \cellcolor[HTML]{C0C0C0}73 &
  \cellcolor[HTML]{C0C0C0}9 &
  \cellcolor[HTML]{C0C0C0}16 &
  \cellcolor[HTML]{C0C0C0}30 &
  \cellcolor[HTML]{EFEFEF}45 &
  \cellcolor[HTML]{C0C0C0}6 \\
 &
  Fax-AI &
  \cellcolor[HTML]{C0C0C0}73 &
  \cellcolor[HTML]{C0C0C0}70 &
  \cellcolor[HTML]{EFEFEF}72 &
  \cellcolor[HTML]{EFEFEF}71 &
  \cellcolor[HTML]{C0C0C0}10 &
  \cellcolor[HTML]{C0C0C0}14 &
  \cellcolor[HTML]{EFEFEF}28 &
  \cellcolor[HTML]{EFEFEF}44 &
  \cellcolor[HTML]{9B9B9B}0 \\
 &
  MAAT &
  \cellcolor[HTML]{C0C0C0}72 &
  \cellcolor[HTML]{9B9B9B}72 &
  \cellcolor[HTML]{EFEFEF}73 &
  \cellcolor[HTML]{EFEFEF}72 &
  \cellcolor[HTML]{EFEFEF}19 &
  \cellcolor[HTML]{C0C0C0}13 &
  \cellcolor[HTML]{C0C0C0}24 &
  \cellcolor[HTML]{C0C0C0}36 &
  \cellcolor[HTML]{C0C0C0}8 \\
 &
  FairMask &
  \cellcolor[HTML]{9B9B9B}74 &
  \cellcolor[HTML]{9B9B9B}71 &
  \cellcolor[HTML]{C0C0C0}75 &
  \cellcolor[HTML]{C0C0C0}73 &
  \cellcolor[HTML]{C0C0C0}11 &
  \cellcolor[HTML]{C0C0C0}14 &
  \cellcolor[HTML]{EFEFEF}27 &
  \cellcolor[HTML]{EFEFEF}43 &
  \cellcolor[HTML]{9B9B9B}0 \\
\multirow{-6}{*}{MEPS: Race} &
  GroupCART &
  \cellcolor[HTML]{EFEFEF}70 &
  \cellcolor[HTML]{9B9B9B}71 &
  \cellcolor[HTML]{9B9B9B}96 &
  \cellcolor[HTML]{9B9B9B}82 &
  \cellcolor[HTML]{9B9B9B}2 &
  \cellcolor[HTML]{9B9B9B}3 &
  \cellcolor[HTML]{9B9B9B}5 &
  \cellcolor[HTML]{9B9B9B}5 &
  \cellcolor[HTML]{9B9B9B}0 \\ \hline
\end{tabular}\label{rq3-1}
}
\end{table*}

\begin{table*}[t!]
\centering
\caption{Results for RQ3. The table shows the mean {\it d2h} (``distance to heaven") scores of
Equation~\ref{d2h} across all datasets. The calculation of {\it d2h} scores is described in Eq. \ref{d2h}. Better scores are marked in darker colors. Note that the last line (with GroupCART) often has those
darker (and better) results.}
\resizebox{\textwidth}{!}{%
\begin{tabular}{|c|cccccccc|}
\hline
 &
  Adult: Sex &
  Adult: Race &
  Compas: Sex &
  Compas: Race &
  German: Sex &
  Bank: Age &
  Heart: Age &
  Meps: Race \\ \hline
CART &
  \cellcolor[HTML]{EFEFEF}124 &
  86 &
  \cellcolor[HTML]{C0C0C0}72 &
  \cellcolor[HTML]{C0C0C0}74 &
  \cellcolor[HTML]{EFEFEF}77 &
  122 &
  \cellcolor[HTML]{EFEFEF}113 &
  \cellcolor[HTML]{EFEFEF}79 \\
EGR &
  \cellcolor[HTML]{9B9B9B}85 &
  \cellcolor[HTML]{C0C0C0}76 &
  \cellcolor[HTML]{9B9B9B}62 &
  \cellcolor[HTML]{9B9B9B}66 &
  \cellcolor[HTML]{C0C0C0}53 &
  \cellcolor[HTML]{EFEFEF}100 &
  \cellcolor[HTML]{C0C0C0}102 &
  \cellcolor[HTML]{EFEFEF}79 \\
Fax-AI &
  \cellcolor[HTML]{C0C0C0}90 &
  \cellcolor[HTML]{EFEFEF}81 &
  \cellcolor[HTML]{EFEFEF}74 &
  \cellcolor[HTML]{C0C0C0}74 &
  \cellcolor[HTML]{C0C0C0}54 &
  115 &
  \cellcolor[HTML]{EFEFEF}115 &
  \cellcolor[HTML]{EFEFEF}79 \\
MAAT &
  \cellcolor[HTML]{9B9B9B}85 &
  \cellcolor[HTML]{9B9B9B}60 &
  \cellcolor[HTML]{C0C0C0}72 &
  \cellcolor[HTML]{EFEFEF}79 &
  \cellcolor[HTML]{EFEFEF}73 &
  \cellcolor[HTML]{C0C0C0}90 &
  \cellcolor[HTML]{9B9B9B}74 &
  \cellcolor[HTML]{C0C0C0}74 \\
FairMask &
  \cellcolor[HTML]{C0C0C0}92 &
  \cellcolor[HTML]{C0C0C0}77 &
  \cellcolor[HTML]{EFEFEF}74 &
  \cellcolor[HTML]{C0C0C0}76 &
  \cellcolor[HTML]{C0C0C0}53 &
  108 &
  \cellcolor[HTML]{EFEFEF}112 &
  \cellcolor[HTML]{C0C0C0}76 \\
GroupCART &
  \cellcolor[HTML]{9B9B9B}86 &
  \cellcolor[HTML]{EFEFEF}80 &
  \cellcolor[HTML]{9B9B9B}64 &
  \cellcolor[HTML]{9B9B9B}68 &
  \cellcolor[HTML]{9B9B9B}51 &
  \cellcolor[HTML]{9B9B9B}70 &
  \cellcolor[HTML]{C0C0C0}101 &
  \cellcolor[HTML]{9B9B9B}46 \\ \hline
\end{tabular}%
}
\label{rq3-2}
\end{table*}

\begin{table*}[t!]
\centering
\caption{Result for RQ4. GroupCART is capable of mitigating bias on multiple protected attributes simultaneously. Here cells with significantly better results are marked in a darker color.}
\resizebox{\textwidth}{!}{%
\begin{tabular}{|c|c|c|cccc|ccccc|}
\hline
Dataset &
  Method &
  Protected Attribute &
  Accuracy &
  Precision &
  Recall &
  F1 &
  AOD &
  EOD &
  SPD &
  DI &
  FR \\ \hline
 &
   &
  Sex &
  \cellcolor[HTML]{C0C0C0} &
  \cellcolor[HTML]{C0C0C0} &
  \cellcolor[HTML]{EFEFEF} &
  \cellcolor[HTML]{EFEFEF} &
  13 &
  36 &
  22 &
  90 &
  17 \\
 &
  \multirow{-2}{*}{CART} &
  Race &
  \multirow{-2}{*}{\cellcolor[HTML]{C0C0C0}84} &
  \multirow{-2}{*}{\cellcolor[HTML]{C0C0C0}72} &
  \multirow{-2}{*}{\cellcolor[HTML]{EFEFEF}51} &
  \multirow{-2}{*}{\cellcolor[HTML]{EFEFEF}60} &
  \cellcolor[HTML]{EFEFEF}3 &
  9 &
  9 &
  47 &
  \cellcolor[HTML]{EFEFEF}6 \\ \cline{2-12} 
 &
   &
  Sex &
  \cellcolor[HTML]{C0C0C0} &
  \cellcolor[HTML]{C0C0C0} &
  \cellcolor[HTML]{C0C0C0} &
  \cellcolor[HTML]{C0C0C0} &
  \cellcolor[HTML]{EFEFEF}7 &
  \cellcolor[HTML]{C0C0C0}9 &
  \cellcolor[HTML]{EFEFEF}14 &
  \cellcolor[HTML]{EFEFEF}70 &
  \cellcolor[HTML]{EFEFEF}6 \\
 &
  \multirow{-2}{*}{MAAT} &
  Race &
  \multirow{-2}{*}{\cellcolor[HTML]{C0C0C0}85} &
  \multirow{-2}{*}{\cellcolor[HTML]{C0C0C0}73} &
  \multirow{-2}{*}{\cellcolor[HTML]{C0C0C0}83} &
  \multirow{-2}{*}{\cellcolor[HTML]{C0C0C0}76} &
  \cellcolor[HTML]{EFEFEF}2 &
  \cellcolor[HTML]{C0C0C0}2 &
  \cellcolor[HTML]{C0C0C0}6 &
  \cellcolor[HTML]{C0C0C0}35 &
  \cellcolor[HTML]{EFEFEF}9 \\ \cline{2-12} 
 &
   &
  Sex &
  \cellcolor[HTML]{C0C0C0} &
  \cellcolor[HTML]{EFEFEF} &
  \cellcolor[HTML]{EFEFEF} &
  \cellcolor[HTML]{EFEFEF} &
  \cellcolor[HTML]{C0C0C0}3 &
  \cellcolor[HTML]{C0C0C0}8 &
  \cellcolor[HTML]{C0C0C0}9 &
  \cellcolor[HTML]{C0C0C0}49 &
  \cellcolor[HTML]{C0C0C0}0 \\
\multirow{-6}{*}{Adult} &
  \multirow{-2}{*}{GroupCART} &
  Race &
  \multirow{-2}{*}{\cellcolor[HTML]{C0C0C0}83} &
  \multirow{-2}{*}{\cellcolor[HTML]{EFEFEF}65} &
  \multirow{-2}{*}{\cellcolor[HTML]{EFEFEF}56} &
  \multirow{-2}{*}{\cellcolor[HTML]{EFEFEF}60} &
  \cellcolor[HTML]{C0C0C0}0 &
  \cellcolor[HTML]{C0C0C0}1 &
  \cellcolor[HTML]{C0C0C0}6 &
  \cellcolor[HTML]{C0C0C0}34 &
  \cellcolor[HTML]{C0C0C0}0 \\ \hline
 &
   &
  Sex &
  \cellcolor[HTML]{C0C0C0} &
  \cellcolor[HTML]{C0C0C0} &
  \cellcolor[HTML]{EFEFEF} &
  \cellcolor[HTML]{C0C0C0} &
  \cellcolor[HTML]{EFEFEF}4 &
  \cellcolor[HTML]{EFEFEF}6 &
  \cellcolor[HTML]{EFEFEF}11 &
  \cellcolor[HTML]{EFEFEF}17 &
  22 \\
 &
  \multirow{-2}{*}{CART} &
  Race &
  \multirow{-2}{*}{\cellcolor[HTML]{C0C0C0}65} &
  \multirow{-2}{*}{\cellcolor[HTML]{C0C0C0}67} &
  \multirow{-2}{*}{\cellcolor[HTML]{EFEFEF}69} &
  \multirow{-2}{*}{\cellcolor[HTML]{C0C0C0}68} &
  \cellcolor[HTML]{EFEFEF}3 &
  \cellcolor[HTML]{EFEFEF}10 &
  \cellcolor[HTML]{EFEFEF}14 &
  \cellcolor[HTML]{EFEFEF}22 &
  21 \\ \cline{2-12} 
 &
   &
  Sex &
  \cellcolor[HTML]{C0C0C0} &
  \cellcolor[HTML]{C0C0C0} &
  \cellcolor[HTML]{EFEFEF} &
  \cellcolor[HTML]{EFEFEF} &
  10 &
  \cellcolor[HTML]{EFEFEF}8 &
  \cellcolor[HTML]{EFEFEF}14 &
  \cellcolor[HTML]{EFEFEF}20 &
  \cellcolor[HTML]{EFEFEF}18 \\
 &
  \multirow{-2}{*}{MAAT} &
  Race &
  \multirow{-2}{*}{\cellcolor[HTML]{C0C0C0}67} &
  \multirow{-2}{*}{\cellcolor[HTML]{C0C0C0}66} &
  \multirow{-2}{*}{\cellcolor[HTML]{EFEFEF}66} &
  \multirow{-2}{*}{\cellcolor[HTML]{EFEFEF}66} &
  14 &
  \cellcolor[HTML]{EFEFEF}10 &
  \cellcolor[HTML]{EFEFEF}17 &
  \cellcolor[HTML]{EFEFEF}22 &
  \cellcolor[HTML]{EFEFEF}19 \\ \cline{2-12} 
 &
   &
  Sex &
  \cellcolor[HTML]{EFEFEF} &
  \cellcolor[HTML]{EFEFEF} &
  \cellcolor[HTML]{C0C0C0} &
  \cellcolor[HTML]{C0C0C0} &
  \cellcolor[HTML]{C0C0C0}2 &
  \cellcolor[HTML]{C0C0C0}5 &
  \cellcolor[HTML]{C0C0C0}6 &
  \cellcolor[HTML]{C0C0C0}9 &
  \cellcolor[HTML]{C0C0C0}0 \\
\multirow{-6}{*}{Compas} &
  \multirow{-2}{*}{GroupCART} &
  Race &
  \multirow{-2}{*}{\cellcolor[HTML]{EFEFEF}63} &
  \multirow{-2}{*}{\cellcolor[HTML]{EFEFEF}63} &
  \multirow{-2}{*}{\cellcolor[HTML]{C0C0C0}73} &
  \multirow{-2}{*}{\cellcolor[HTML]{C0C0C0}68} &
  \cellcolor[HTML]{C0C0C0}2 &
  \cellcolor[HTML]{C0C0C0}9 &
  \cellcolor[HTML]{C0C0C0}12 &
  \cellcolor[HTML]{C0C0C0}15 &
  \cellcolor[HTML]{C0C0C0}0 \\ \hline
\end{tabular}%
}
\label{tab:rq4-4}
\end{table*}

\begin{blockquote}
\textbf{RQ2}: How does the size of ensemble learners affect the effectiveness of GroupCART? 
\end{blockquote}
\begin{wrapfigure}{R}{2.5in}
\begin{adjustbox}{max width=.99\textwidth}
\includegraphics[width=.99\linewidth]{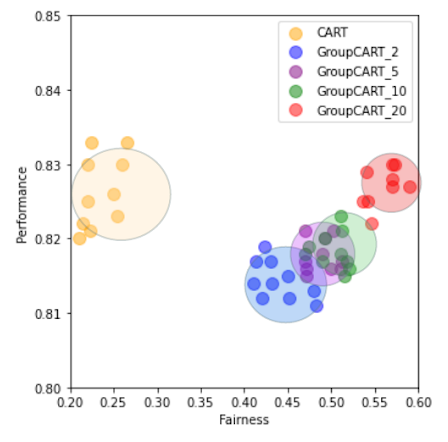}
\end{adjustbox}
\caption{Results for RQ2. Better results have higher fairness and performance (i.e., found top-right).} \label{rq2}
\end{wrapfigure}
Previous work in ensemble fairness-aware learning did not investigate the influence of ensemble sizes on the effectiveness of a model. Therefore, we wonder the question ``Is the size of two large enough for ensemble learning?".
As shown in Figure. \ref{rq2}, we experimented with GroupCART of different ensemble sizes (number of learners based on). 
Recall from \S\ref{rig} that we continued to double the size of the ensembles until the FLIP rates told us that further growth was not required (which meant, in practice, that we used ensembles of size 5, 10, 20). As seen in the Adult case, the quality (as well as variance) of the generated models is better off as the ensemble size increases, showing better stability.
The results imply that, in the case of the Adult dataset, GroupCART can obtain increasingly better trade-offs with the increase of ensemble sizes.

Thus, our answer to RQ2 is: {\bf The experimental results indicate that ensemble learning can benefit from the increase of ensemble sizes. When using GroupCART, better trade-offs are achieved with greater ensemble pools.}

\begin{blockquote}
\textbf{RQ3}: How effective is GroupCART compared to state-of-the-art methods?
\end{blockquote}
To evaluate the overall effectiveness of GroupCART compared to other benchmark methods, we measure a model in three aspects: predictive performance, group fairness, and individual fairness. Prediction performance is measured using the four binary classification metrics described in  Table~\ref{tab:metrics2}. As for group and individual fairness, we use five different metrics as described in Table~\ref{tab:metrics}.

We present the comprehensive comparison in Table~\ref{rq3-1}. The table shows that GroupCART constantly obtained on-par or superior scores across different metrics. It should also be noted that GroupCART always obtains much better FR (flip rate) scores, indicating that GroupCART can usually ensure perfect individual fairness.  
Table~\ref{rq3-2} shows the {\it d2h} scores computed using the nine metrics of Table~\ref{rq3-1}. The {\it d2h} scores provide a generalized summary regarding the trade-offs between performance and fairness. As seen in the table, GroupCART is constantly ranked first place in most cases (6 out of 8). In the rest of the two cases, GroupCART also achieved better {\it d2h} scores than the baseline CART model.
Thus, our answer to RQ3 is: {\bf GroupCART performs better or similar to other state-of-the-art methods in terms of performance and fairness.}

\begin{blockquote}
\textbf{RQ4}: How effective is GroupCART when dealing with multiple protected attributes?
\end{blockquote}

\begin{figure*}[t!]
     \centering
     \begin{subfigure}[b]{0.3\textwidth}
         \centering
         \includegraphics[width=\textwidth]{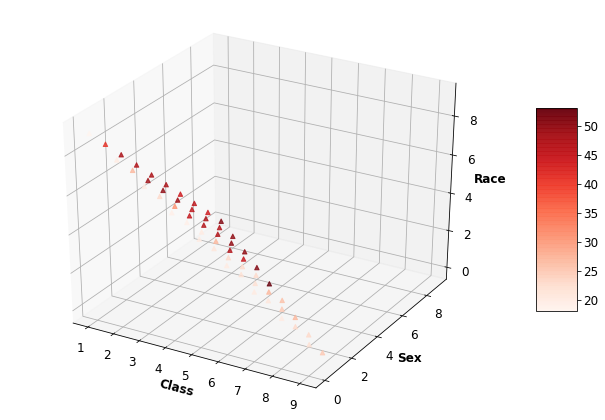}
         \caption{DI scores of the protected attribute {\it sex} achieved by GroupCART in the Adult dataset.\\}
         \label{fig:rq4-1}
     \end{subfigure}
     \hfill
     \begin{subfigure}[b]{0.3\textwidth}
         \centering
         \includegraphics[width=\textwidth]{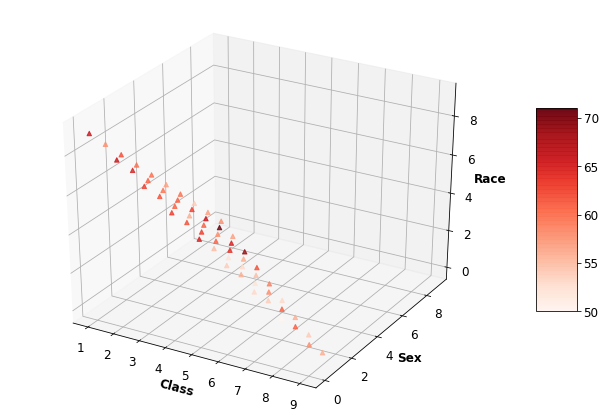}
         \caption{DI scores of the protected attribute {\it race} achieved by GroupCART in the Adult dataset.}
         \label{fig:rq4-2}
     \end{subfigure}
     \hfill
     \begin{subfigure}[b]{0.3\textwidth}
         \centering
         \includegraphics[width=\textwidth]{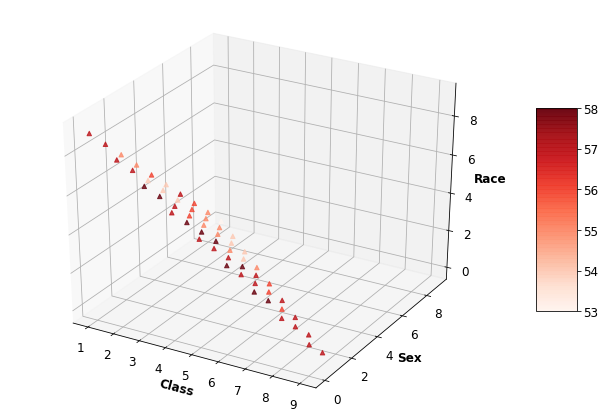}
         \caption{F1 scores achieved by GroupCART in the Adult dataset.\\ {  }}
         \label{fig:rq4-3}
     \end{subfigure}
        \caption{Results for RQ4. The two figures on the left-hand side present disparate impact (DI) scores on two different protected attributes in the Adult dataset, and darker colors indicate better fairness.}
        \label{fig:rq4}
\end{figure*}

The algorithmic designs of GroupCART make it applicable to handle multiple protected attributes simultaneously. As demonstrated in Figure. \ref{fig:rq4}, in the case of the Adult dataset, GroupCART can provide ``smooth performance-fairness trade-offs when tuning internal hyperparameter configurations. With different weights being assigned to IGC and IGS on different protected attributes (in this case, sex and race), the GroupCART model can be guided to concentrate on either the performance or fairness of certain protected attributes. 
We compare GroupCART with MAAT, which is also designed to handle multiple protected attributes. As shown in Table~\ref{tab:rq4-4}, GroupCART can provide better trade-offs in the Compas dataset with respect to both protected attributes. In the Adult dataset, GroupCART provides better fairness yet exhibits a slight compromise in performance.

Thus, our answer to RQ4 is as follows: {\bf GroupCART performs better or similar to the baseline methods when handling multiple (two) protected attributes.}

\section{Threats to Validity}
\label{threat}

\noindent
\textbf{Sampling Bias} - While experimenting with other datasets may yield different results, we believe that our extensive study here has shown the constant effectiveness of GroupCART in various cases. Most of the prior works~\cite{NIPS2017_6988,Galhotra_2017,zhang2018mitigating,Kamiran:2018:ERO:3165328.3165686,chakraborty2019software} used one or two datasets where we used seven well-known datasets in our experiments. We have also observed other emerging datasets in the fairness fields, and we will try to extend our research scope once we verify the validity of the new datasets. In the future, we will explore more datasets and more learners. \\
\textbf{Evaluation Bias} - We used the five fairness metrics in this study, covering both the definitions of group and individual fairness. Previous work \cite{Chakraborty_2020,hardt2016equality,10.1007/978-3-642-33486-3_3} used fewer metrics whereas IBM AIF360~\cite{bellamy2018ai} contains more than 50 metrics. More evaluation criteria will be examined in future work.\\
\textbf{Conclusion Validity} - Our experiments are based on the assumption that the test data are unbiased and correctly labeled. Prior fairness studies also made a similar assumption~\cite{Chakraborty2021BiasIM,Biswas_2020,chakraborty2019software}.\\
\textbf{Internal Validity} - When GroupCART performs multi-objective optimization, we chose four metrics as objective scores (two for performance, two for fairness) for the calculation of continuous domination. This design choice is based on preliminary experiments where we observe that different configurations have greater variance in these metrics. However, it remains possible that an alternative combination of different/more metrics may yield better results. In future work, we will explore more combinations of objective scores for a more comprehensive comparison study. \\
\textbf{Instrumental Validity} The optimization approach in this paper specifically relies on the fairness-aware tree-based learning algorithm. While our approach currently only applies to tree-based algorithms such as random forest or gradient-boosted trees, we believe the idea of using the Pareto Frontier to optimize the fairness-performance trade-offs can be extended to all kinds of classification models.\\
\textbf{External Validity} - Our work is limited to binary classification and tabular data which are very common in AI software. However, all the methods used in this paper can easily be extended in the case of multi-class classification and regression problems. In the future, we will try to extend our work to other domains of SE and ML.

\section{Conclusion}
Ethical concerns rise as an increasingly important topic in the field of ML software. It is widely believed in the SE community that unfairness is a kind of software bug that requires systemic prevention and fixing. In this paper, we attempt to use domain knowledge from the software configuration field to address this problem. That is, we believe that performance-fairness trade-offs can be treated as a software configuration optimization problem. In this spirit, we imported methodologies such as the Pareto frontier filtering to select a subset of models that provide various optimums in the performance-fairness dilemma.

By treating models with different concentrations (as controlled by the internal weighting mechanism) as different hyperparameter configurations, we found that (a) we can usually find multiple trade-off optimums given some performance and fairness metrics, and (b) we can provide better and fairer predictions when aggregating multiple optimal models. Compared to prior works in ensemble fairness-aware learning, MAAT, our approach can provide extensively more options to tune a model along the spectrum of performance-fairness trade-off. Moreover, by enabling the users to customize the importance weights on performance and fairness, our approach is more flexible and transparent in the manner that users can perceive how their preferences/requirements influence the shift of a model's behavior (as illustrated by the non-dominated solutions in Figure. \ref{pareto}).
In summary, we conclude that:
\begin{itemize}
    \item We can treat fairness-aware learning as a software configuration optimization problem.
    \item We can use prominent optimization heuristics to obtain on-par or superior trade-offs compared to benchmark methods.
    \item We can recommend using GroupCART if one wants to build customized models with specific constraints on performance and fairness.
\end{itemize}

\section*{Acknowledgements}
This work was partially funded by 
a research grant from the Laboratory for Analytical Sciences, North Carolina State University.

\section*{Declarations}
\subsection*{Funding and Conflicts of interests}
Apart from the funding acknowledged above, this work does not have any other conflicts of interests.



\bibliographystyle{spbasic}      
\bibliography{main}              

\end{document}